\documentclass[10pt,twocolumn,letterpaper]{article}

\usepackage{cvpr}
\usepackage{times}
\pdfoutput=1
\usepackage{epsfig}
\usepackage{amsmath}
\usepackage{amssymb}
\usepackage{booktabs}
\usepackage{pifont}
\usepackage{float}
\usepackage{cuted}
\usepackage{capt-of}
\usepackage{subcaption}
\usepackage{graphicx} 
\usepackage[hyphens]{url}

\usepackage{amsmath}
\DeclareMathOperator*{\argmax}{arg\,max}

\usepackage[breaklinks=False,bookmarks=false]{hyperref}

\usepackage{paralist}

\usepackage{caption}
\usepackage{subcaption}
\usepackage{float}
\cvprfinalcopy 

\def\cvprPaperID{4781} 

\ifcvprfinal\fi
\begin{document}

\title{Inverse Cooking: Recipe Generation from Food Images}

\author{Amaia Salvador$^1$\thanks{Work done during internship at Facebook AI Research} \: \:\: Michal Drozdzal$^2$ \: Xavier Giro-i-Nieto$^1$ \: Adriana Romero$^2$ \\
\and
$^1$Universitat Politecnica de Catalunya \:\:\:
$^2$Facebook AI Research \\
\small{\{amaia.salvador, xavier.giro\}@upc.edu, \{adrianars, mdrozdzal\}@fb.com} }
\maketitle

\begin{abstract}
People enjoy food photography because they appreciate food. Behind each meal there is a story described in a complex recipe and, unfortunately, by simply looking at a food image we do not have access to its preparation process. Therefore, in this paper we introduce an inverse cooking system that recreates cooking recipes given food images. Our system predicts ingredients as sets by means of a novel architecture, modeling their dependencies without imposing any order, and then generates cooking instructions by attending to both image and its inferred ingredients simultaneously. We extensively evaluate the whole system on the large-scale Recipe1M dataset and show that (1) we improve performance w.r.t. previous baselines for ingredient prediction; (2) we are able to obtain high quality recipes by leveraging both image and ingredients; (3) our system is able to produce more compelling recipes than retrieval-based approaches according to human judgment. We make code and models publicly available\footnote{\scriptsize \url{https://github.com/facebookresearch/inversecooking}}.

\end{abstract}

\section{Introduction}
\label{sec_intro}
Food is fundamental to human existence. Not only does it provide us with energy---it also defines our identity and culture \cite{Fischler1988, min2018you}. As the old saying goes, \emph{we are what we eat}, and food related activities such as cooking, eating and talking about it take a significant portion of our daily life. Food culture has been spreading more than ever in the current digital era, with many people sharing pictures of food they are eating across social media~\cite{instagramfood}. Querying Instagram for \#food leads to at least 300M posts; similarly, searching for \#foodie results in at least 100M posts, highlighting the unquestionable value that food has in our society. Moreover, eating patterns and cooking culture have been evolving over time. In the past, food was mostly prepared at home, but nowadays we frequently consume food prepared by third-parties (e.g. takeaways, catering and restaurants). Thus, the access to detailed information about prepared food is limited and, as a consequence, it is hard to know precisely what we eat. Therefore, we argue that there is a need for \emph{inverse cooking} systems, which are able to infer ingredients and cooking instructions from a prepared meal. 

The last few years have witnessed outstanding improvements in visual recognition tasks such as natural image classification \cite{SimonyanZ14a, HeZR015}, object detection \cite{RenHG015,RedmonDGF15} and semantic segmentation \cite{long2015fully, jegou2017one}. However, when comparing to natural image understanding, food recognition poses additional challenges, since food and its components have high intra-class variability and present heavy deformations that occur during the cooking process. Ingredients are frequently occluded in a cooked dish and come in a variety of colors, forms and textures. Further, visual ingredient detection requires high level reasoning and prior knowledge (e.g. \emph{cake} will likely contain \emph{sugar} and not \emph{salt}, while \emph{croissant} will presumably include \emph{butter}). Hence, food recognition challenges current computer vision systems to go beyond the merely visible, and to incorporate prior knowledge to enable high-quality structured food preparation descriptions.

\begin{figure}
  \includegraphics[width=\columnwidth]{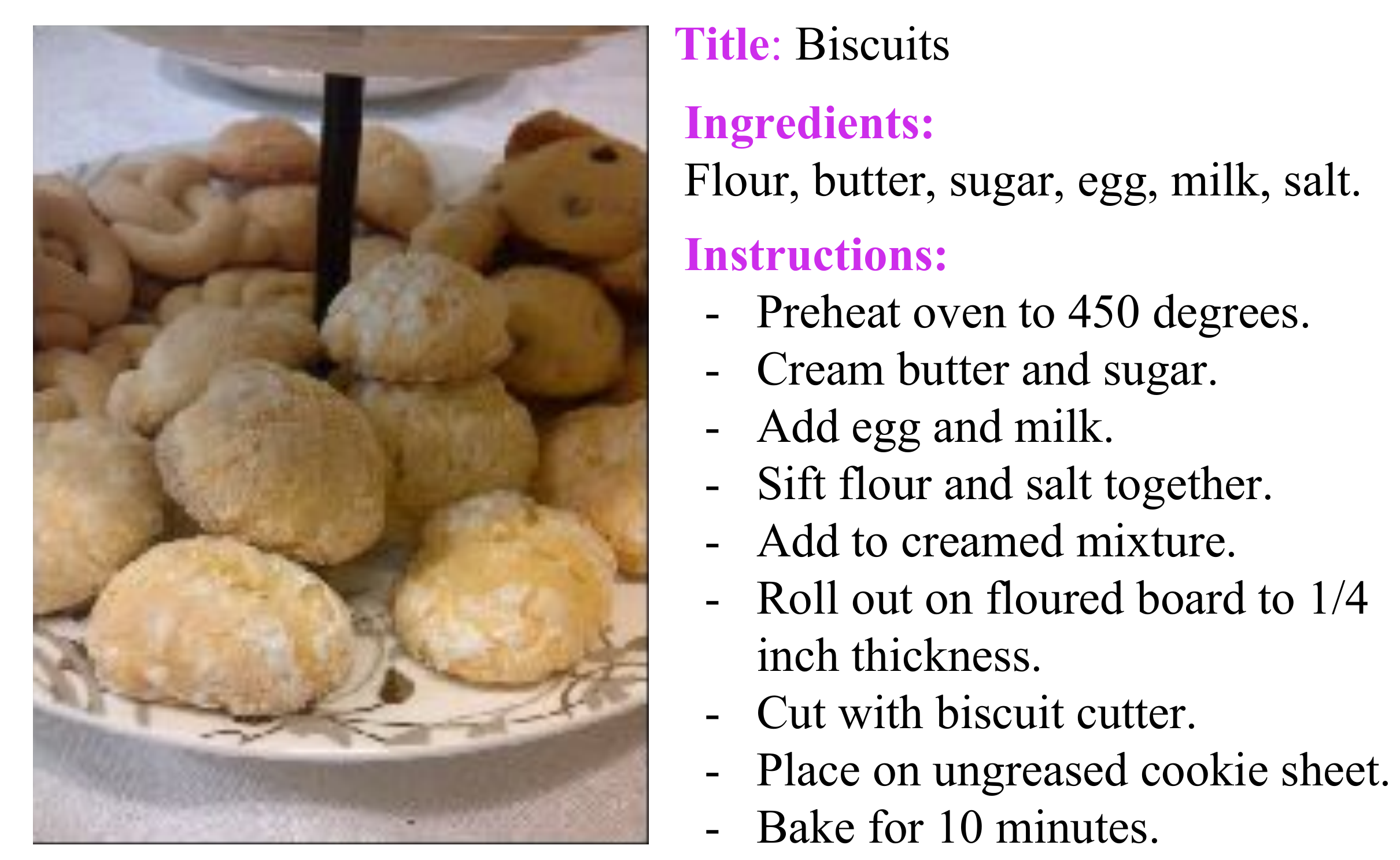}
  \caption{\textbf{Example of a generated recipe}, composed of a title, ingredients and cooking instructions.}
  \label{fig:intro_fig}
  \vspace{-3mm}
\end{figure}

Previous efforts on food understanding have mainly focused on food and ingredient categorization \cite{food101, ofli2017saki, Lee_2018_CVPR}. However, a system for comprehensive visual food recognition should not only be able to recognize the type of meal or its ingredients, but also understand its preparation process. Traditionally, the image-to-recipe problem has been formulated as a retrieval task \cite{wang2015recipe,chen2016deep, chen2017cross, recipe1m}, where a recipe is retrieved from a fixed dataset based on the image similarity score in an embedding space. The performance of such systems highly depends on the dataset size and diversity, as well as on the quality of the learned embedding. Not surprisingly, these systems fail when a matching recipe for the image query does not exist in the static dataset. 

An alternative to overcome the dataset constraints of retrieval systems is to formulate the image-to-recipe problem as a conditional generation one. Therefore, in this paper, we present a system that \emph{generates} a cooking recipe containing a title, ingredients and cooking instructions directly from an image. Figure \ref{fig:intro_fig} shows an example of a generated recipe obtained with our method, which first predicts ingredients from an image and then conditions on both the image and the ingredients to generate the cooking instructions. To the best of our knowledge, our system is the first to \emph{generate} cooking recipes directly from food images. We pose the instruction generation problem as a sequence generation one \emph{conditioned on two modalities} simultaneously, namely an image and its predicted ingredients. We formulate the ingredient prediction problem as a \emph{set prediction}, exploiting their underlying structure. We model ingredient dependencies while not penalizing for prediction order, thus revising the question of \emph{whether order matters} \cite{Vinyals16ordermatters}. We extensively evaluate our system on the large-scale Recipe1M dataset \cite{recipe1m} that contains images, ingredients and cooking instructions, showing satisfactory results. More precisely, in a human evaluation study, we show that our inverse cooking system outperforms previously introduced image-to-recipe retrieval approaches by a large margin. Moreover, using a small set of images, we show that food image-to-ingredient prediction is a hard task for humans and that our approach is able to surpass them.

The contributions of this paper can be summarized as:
\begin{compactitem}
    \item[--] We present an inverse cooking system, which generates cooking instructions conditioned on an image and its ingredients, exploring different attention strategies to reason about both modalities simultaneously.
    \item[--] We exhaustively study ingredients as both a \emph{list} and a \emph{set}, and propose a new architecture for ingredient prediction that exploits co-dependencies among ingredients without imposing order.
    \item[--] By means of a user study we show that ingredient prediction is indeed a difficult task and demonstrate the superiority of our proposed system against image-to-recipe retrieval approaches.
\end{compactitem}

\section{Related Work}
\label{sec_related}
\begin{figure*}
  \includegraphics[width=\textwidth]{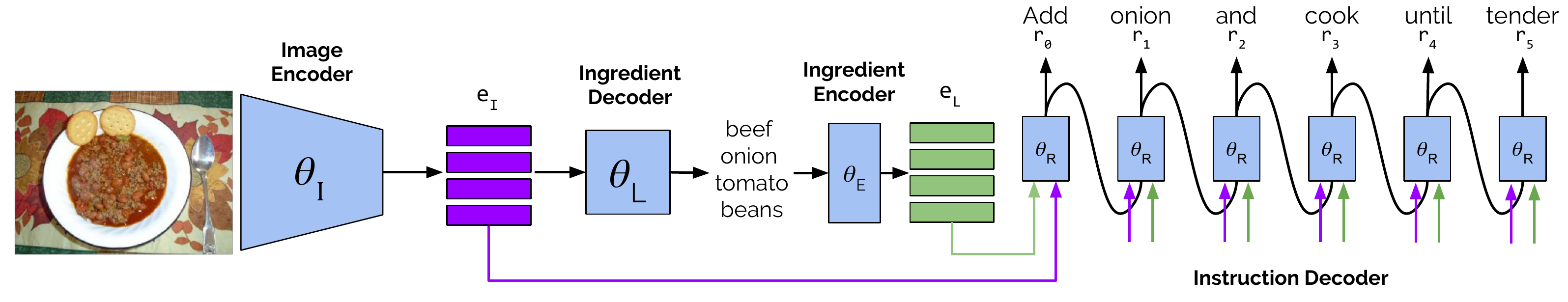}
  \caption{\textbf{Recipe generation model.} We extract image features $\mathbf{e}_I$ with the image encoder, parametrized by $\theta_{I}$. Ingredients are predicted by $\theta_{L}$, and encoded into ingredient embeddings $\mathbf{e}_{L}$ with $\theta_{e}$. The cooking instruction decoder, parametrized by $\theta_{R}$ generates a recipe title and a sequence of cooking steps by attending to image embeddings $\mathbf{e}_{I}$, ingredient embeddings $\mathbf{e}_{L}$, and previously predicted words $(r_{0}, ..., r_{t-1})$. }
  \label{fig:full_model}
  \vspace{-3mm}
\end{figure*}
\textbf{Food Understanding.} The introduction of large scale food datasets, such as Food-101 \cite{food101} and Recipe1M \cite{recipe1m}, together with a recently held iFood challenge\footnote{\url{https://www.kaggle.com/c/ifood2018}} has enabled significant advancements in visual food recognition, by providing reference benchmarks to train and compare machine learning approaches. As a result, there is currently a vast literature in computer vision dealing with a variety of food related tasks, with special focus in image classification \cite{Liu2016DDL,ofli2017saki,Ngo2017DLF,nu9070657,chen2017chinesefoodnet,Lee_2018_CVPR, martinel2018wide, xu2015geolocalized, herranz2017modeling, horiguchi2018personalized}. 
Subsequent works tackle more challenging tasks such as estimating the number of calories given a food image \cite{im2calories}, estimating food quantities \cite{Chen2012quantities}, predicting the list of present ingredients \cite{chen2016deep, chen2017cross} and finding the recipe for a given image \cite{wang2015recipe,chen2016deep, chen2017cross, recipe1m, carvalho2018cross}. Additionally, \cite{min2018you} provides a detailed cross-region analysis of food recipes, considering images, attributes (e.g. style and course) and recipe ingredients. Food related tasks have also been considered in the natural language processing literature, where recipe generation has been studied in the context of generating procedural text from either flow graphs \cite{Hammond86,MoriMYS14,MoriMSYHFY14} or ingredients' checklists \cite{Kiddon16}.

\textbf{Multi-label classification.} Significant effort has been devoted in the literature to leverage deep neural networks for multi-label classification, by designing models \cite{Tsoumakas10powerset,Dembczynski2010pcc,WeiXHNDZY14,Nam17setrnn,WangYMHHX16cnnrnn} and studying loss functions \cite{GongJLTI13} well suited for this task. Early attempts exploit single-label classification models coupled with binary logistic loss \cite{chen2016deep}, assuming the independence among labels and dropping potentially relevant information. One way of capturing label dependencies is by relying on label powersets \cite{Tsoumakas10powerset}. Powersets consider all possible label combinations, which makes them intractable for large scale problems. Another expensive alternative consists in learning the joint probability of the labels. To overcome this issue, probabilistic classifier chains \cite{Dembczynski2010pcc} and their recurrent neural network-based \cite{WangYMHHX16cnnrnn,Nam17setrnn} counterparts propose to decompose the joint distribution into conditionals, at the expense of introducing intrinsic ordering. Note that most of these models require to make a prediction for each of the potential labels. Moreover, joint input and label embeddings \cite{Weston2011WSU,Lin2014MCV,YehWKW17} have been introduced to preserve correlations and predict label sets. As an alternative, researchers have attempted to predict the cardinality of the set of labels \cite{deepsetnet,rezatofighi2017joint}; however, assuming the independence of labels. When it comes to multi-label classification objectives, binary logistic loss \cite{chen2016deep}, target distribution cross-entropy \cite{GongJLTI13,Mahajan18hashtags}, target distribution mean squared error \cite{WeiXHNDZY14} and ranking-based losses \cite{GongJLTI13} have been investigated and compared. Recent results on large scale datasets outline the potential of the target distribution loss \cite{Mahajan18hashtags}.
 
\textbf{Conditional text generation.} Conditional text generation with auto-regressive models has been widely studied in the literature using both text-based \cite{seq2seq, convs2s, transformer, fan2018hierarchical} as well as image-based conditionings \cite{showtell, show_attend_tell, sentinel, Karpathy:2017, krause2017hierarchical,Dai2017TowardsDA,sharma2018conceptual}. In neural machine translation, where the goal is to predict the translation for a given source text into another language, different architecture designs have been studied, including recurrent neural networks \cite{seq2seq}, convolutional models \cite{convs2s} and attention based approaches \cite{transformer}. More recently, sequence-to-sequence models have been applied to more open-ended generation tasks, such as poetry \cite{WangHWWLWC16} and story generation \cite{krause2017hierarchical,fan2018hierarchical}. Following neural machine translation trends, auto-regressive models have exhibited promising performance in image captioning \cite{showtell, show_attend_tell, sentinel, Karpathy:2017,Dai2017TowardsDA,sharma2018conceptual}, where the goal is to provide a short description of the image contents, opening the doors to less constrained problems such as generating descriptive paragraphs \cite{krause2017hierarchical} or visual storytelling \cite{Huang2018}.

\section{Generating recipes from images}
\label{sec_method}
\begin{figure*}[t]
    \centering
    \begin{subfigure}[t]{0.24\textwidth}
        \includegraphics[width=\textwidth]{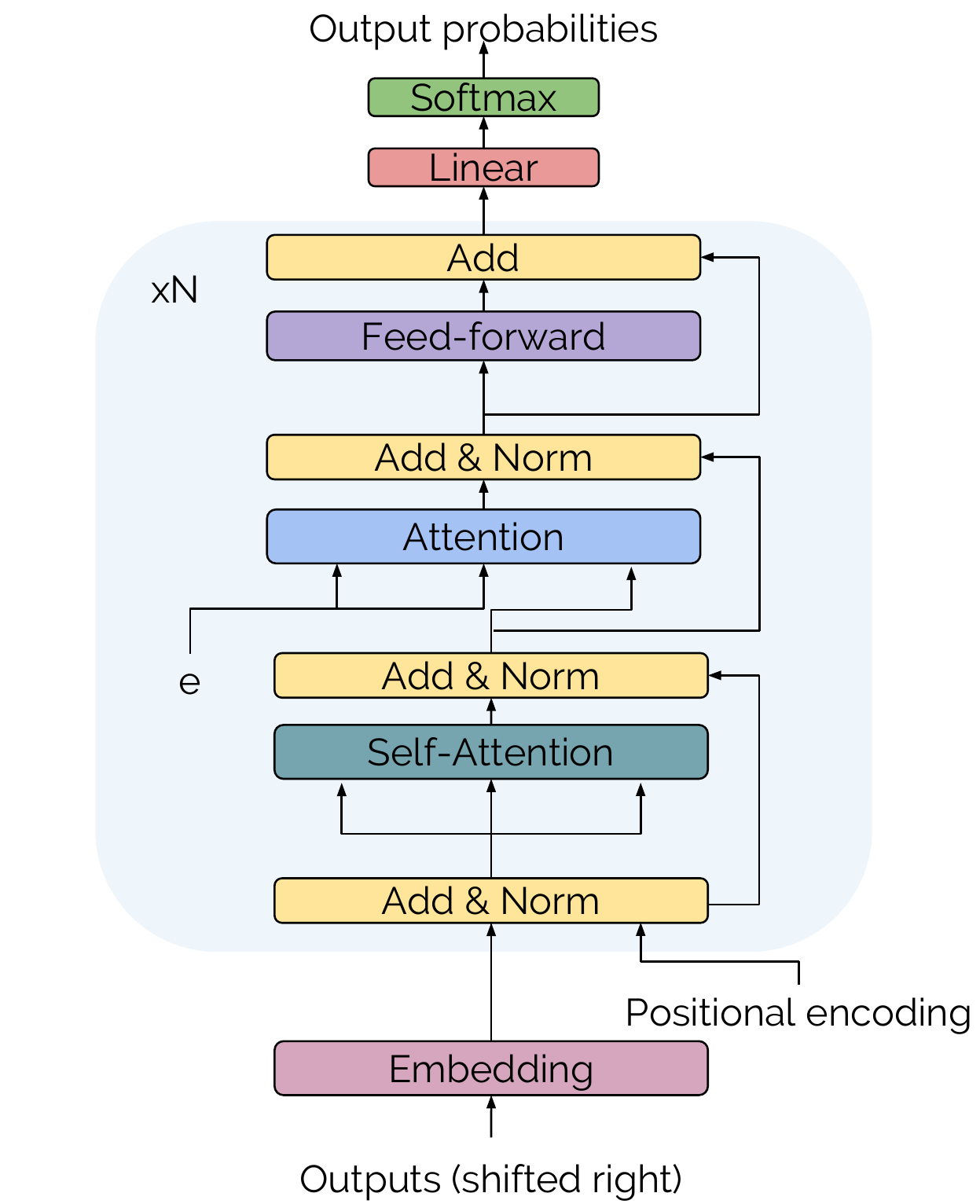}
        \caption{Transformer model \cite{transformer}}
        \label{fig:tf_original}
    \end{subfigure}
    \vline
    \begin{subfigure}[t]{0.24\textwidth}
        \includegraphics[width=\textwidth]{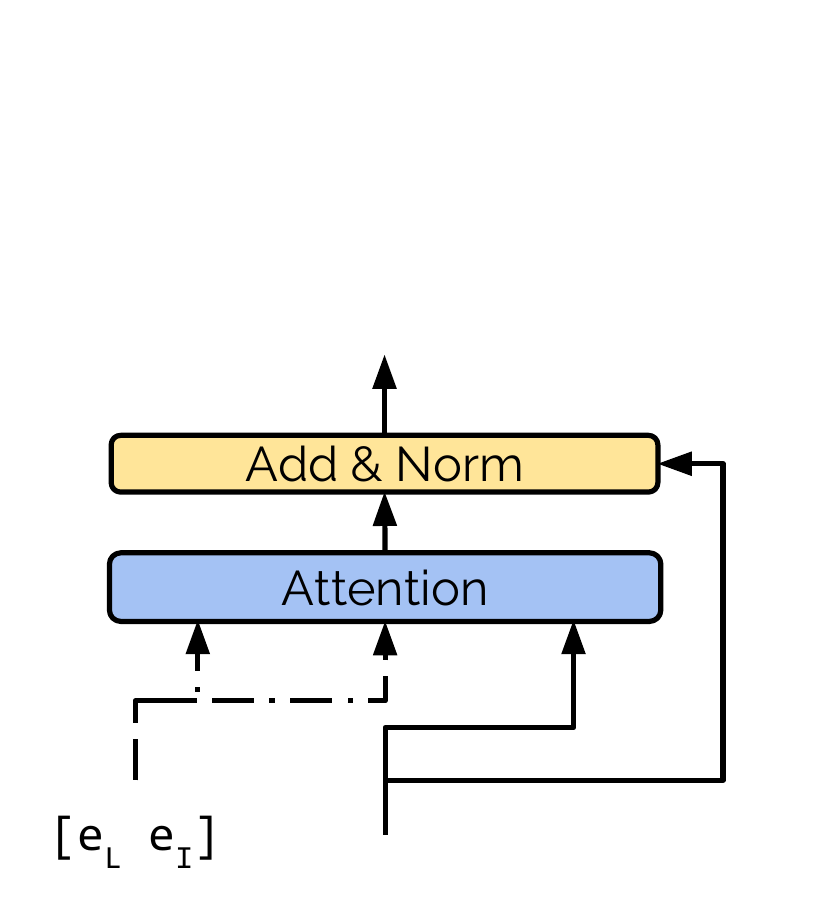}
        \caption{Concatenated}
        \label{fig:tf_concat}
    \end{subfigure}
        \begin{subfigure}[t]{0.24\textwidth}
        \includegraphics[width=\textwidth]{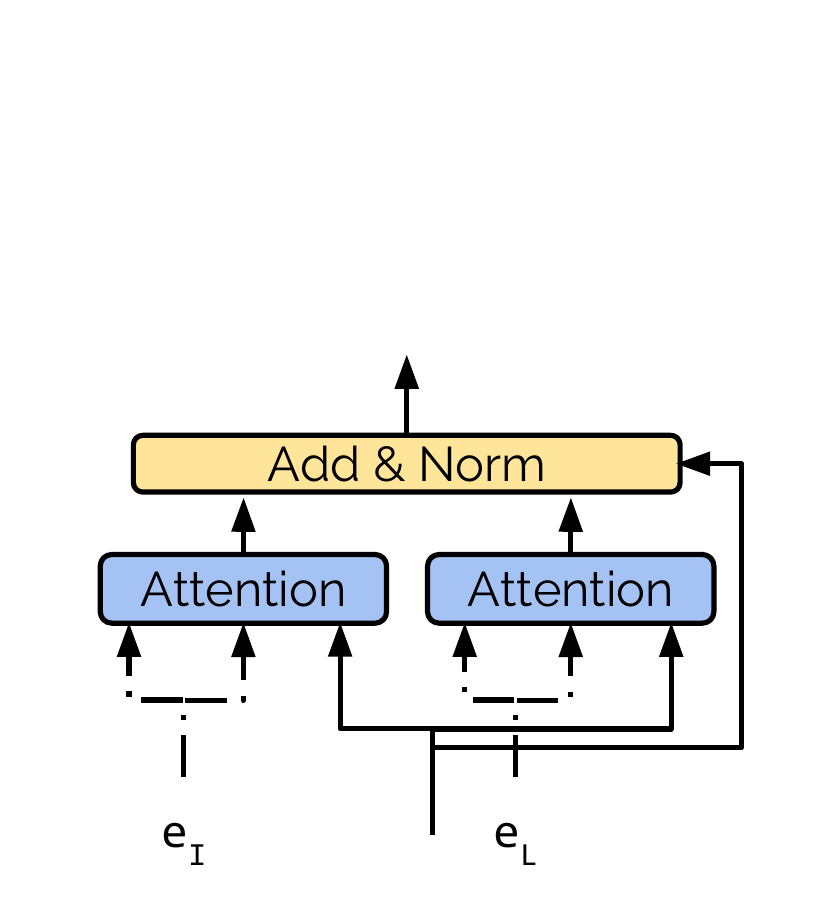}
        \caption{Independent}
        \label{fig:tf_independent}
    \end{subfigure}
    \begin{subfigure}[t]{0.24\textwidth}
        \includegraphics[width=\textwidth]{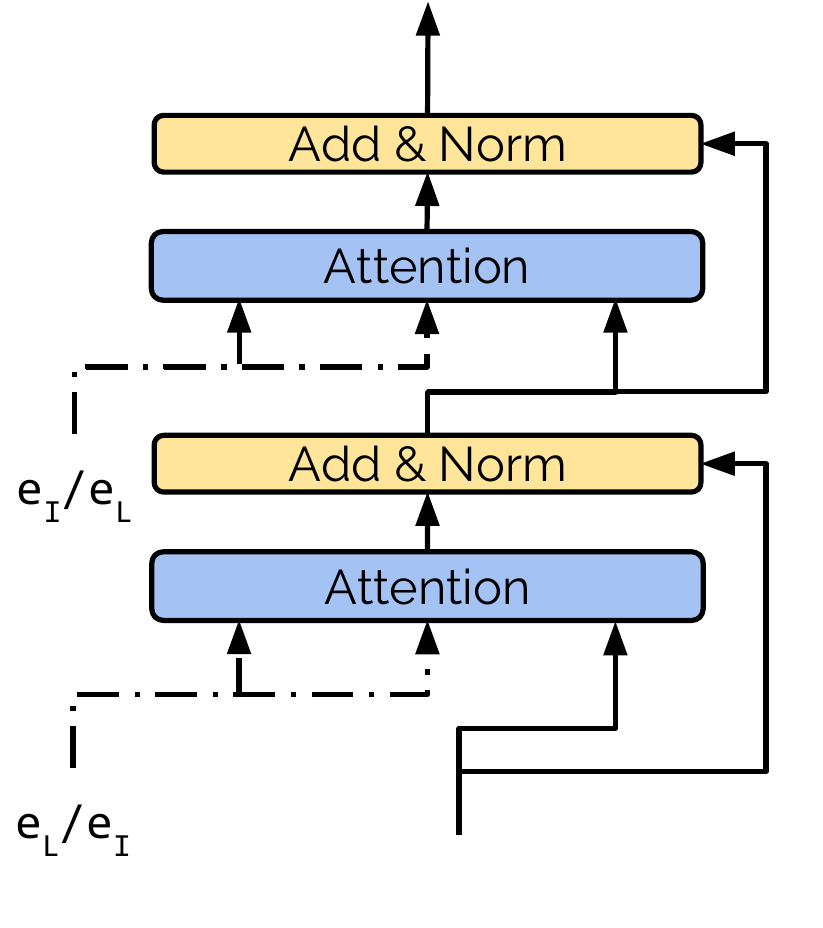}
        \caption{Sequential}
        \label{fig:tf_seq}
    \end{subfigure}
    \caption{\textbf{Attention strategies for the instruction decoder.} In our experiments, we replace the attention module in the transformer (a), with three different attention modules (b-d) for cooking instruction generation using multiple conditions.}\label{fig:att_types}
    \vspace{-3mm}
\end{figure*}
Generating a recipe (title, ingredients and instructions) from an image is a challenging task, which requires a simultaneous understanding of the ingredients composing the dish as well as the transformations they went through, e.g. slicing, blending or mixing with other ingredients. Instead of obtaining the recipe from an image directly, we argue that a recipe generation pipeline would benefit from an intermediate step predicting the ingredients list. The sequence of instructions would then be generated conditioned on both the image and its corresponding list of ingredients, where the interplay between image and ingredients could provide additional insights on how the latter were processed to produce the resulting dish. 

Figure \ref{fig:full_model} illustrates our approach. Our recipe generation system takes a food image as an input and outputs a sequence of cooking instructions, which are generated by means of an instruction decoder that takes as input two embeddings. The first one represents visual features extracted from an image, while the second one encodes the ingredients extracted from the image. We start by introducing our transfomer-based instruction decoder in Subsection \ref{sec_recipegen}. This allows us to formally review the transformer, which we then study and modify to predict ingredients in an orderless manner in Subsection \ref{sec_ingrpred}. Finally, we review the optimization details in Subsection \ref{sec_optim}.

\subsection{Cooking Instruction Transformer}
\label{sec_recipegen}
Given an input image with associated ingredients, we aim to produce a sequence of instructions $R = (r_1,..., r_T)$ (where $r_t$ denotes a word in the sequence) by means of an instruction transformer \cite{transformer}. Note that the title is predicted as the first instruction. This transformer is conditioned jointly on two inputs: the image representation $\mathbf{e}_I$ and the ingredient embedding $\mathbf{e}_L$. We extract the image representation with a ResNet-50 \cite{resnet} encoder and obtain the ingredient embedding $\mathbf{e}_L$ by means of a decoder architecture to predict ingredients, followed by a single embedding layer mapping each ingredient into a fixed-size vector. 

The instruction decoder is composed of \emph{transformer blocks}, each of them containing two attention layers followed by a linear layer \cite{transformer}. The first attention layer applies self-attention over previously generated outputs, whereas the second one attends to the model conditioning in order to refine the self-attention output. 
The transformer model is composed of multiple transformer blocks followed by a linear layer and a softmax nonlinearity that provides a distribution over recipe words for each time step $t$. Figure \ref{fig:tf_original} illustrates the transformer model, which traditionally is conditioned on a single modality. However, our recipe generator is conditioned on two sources: the image features $\mathbf{e}_I \in \mathbb{R}^{P\times d_{e}}$ and ingredients embeddings $\mathbf{e}_L \in \mathbb{R}^{K\times d_{e}}$ ($P$ and $K$ denote the number of image and ingredient features, respectively, and $d_{e}$ is the embedding dimensionality). Thus, we want our attention to reason about both modalities simultaneously, guiding the instruction generation process. To that end, we explore three different fusion strategies (depicted in Figure \ref{fig:att_types}):
\begin{compactitem}
\item[--]\textbf{Concatenated attention.} This strategy first concatenates both image $\mathbf{e}_I$ and ingredients $\mathbf{e}_L$ embeddings over the first dimension $\mathbf{e}_{concat} \in \mathbb{R}^{(K+P)\times d_{e}}$. Then, attention is applied over the combined embeddings. 
\item[--]\textbf{Independent attention.} This strategy incorporates two attention layers to deal with the bi-modal conditioning. In this case, one layer attends over the image embedding $\mathbf{e}_I$, whereas the other attends over the ingredient embeddings $\mathbf{e}_L$. The output of both attention layers is combined via summation operation. 
\item[--]\textbf{Sequential attention.} This strategy sequentially attends over the two conditioning modalities. In our design, we consider two orderings: (1) \emph{image first} where the attention is first computed over image embeddings $\mathbf{e}_I$ and then over ingredient embeddings $\mathbf{e}_L$; and (2) \emph{ingredients first} where the order is flipped and we first attend over ingredient embeddings $\mathbf{e}_L$ followed by image embeddings $\mathbf{e}_I$.  
\end{compactitem}

\subsection{Ingredient Decoder}
\label{sec_ingrpred}

Which is the best structure to represent ingredients? On the one hand, it seems clear that ingredients are a \emph{set}, since permuting them does not alter the outcome of the cooking recipe. On the other hand, we colloquially refer to ingredients as a \emph{list} (e.g. list of ingredients), implying some order. Moreover, it would be reasonable to think that there is some information in the order in which humans write down the ingredients in a recipe. Therefore, in this subsection we consider both scenarios and introduce models that work either with a list of ingredients or with a set of ingredients. 

A \emph{list of ingredients} is a variable sized, ordered collection of unique meal constituents. More precisely, let us define a dictionary of ingredients of size $N$ as $\mathcal{D} = \{d_i\}_{i=0}^{N}$, from which we can obtain a list of ingredients $L$ by selecting $K$ elements from $\mathcal{D}$: $L = [l_i]_{i=0}^{K}$. We encode $L$ as a binary matrix $\mathbf{L}$ of dimensions $K \times N$, with  $\mathbf{L}_{i,j}=1$ if $d_j \in \mathcal{D}$ is selected and $0$ otherwise (one-hot-code representation). 
Thus, our training data consists of $M$ image and ingredient list pairs $\{(\mathbf{x}^{(i)}, \mathbf{L}^{(i)})\}_{i=0}^{M}$. In this scenario, the goal is to predict $\mathbf{\hat{L}}$ from an image $\mathbf{x}$ by maximizing the following objective:
\begin{equation}
   \argmax_{\theta_I, \theta_L} \sum_{i=0}^M \log p(\mathbf{\hat{L}}^{(i)}=\mathbf{L}^{(i)}|\mathbf{x}^{(i)};\theta_I, \theta_L),
\label{eq:ing_list}
\end{equation}
where $\theta_I$ and $\theta_L$ represent the learnable parameters of the image encoder and ingredient decoder, respectively. Since $\mathbf{L}$ denotes a list, we can factorize $p(\mathbf{\hat{L}}^{(i)}=\mathbf{L}^{(i)}|\mathbf{x}^{(i)})$ into $K$ conditionals:
$\sum_{k=0}^K \log p(\mathbf{\hat{L}}_{k}^{(i)}=\mathbf{L}_{k}^{(i)}|\mathbf{x}^{(i)}, \mathbf{L}_{<k}^{(i)})$~\footnote{ $\mathbf{L}_{k}^{(i)}$ denotes the $k$-th row of $\mathbf{L}^{(i)}$ and $\mathbf{L}_{<k}^{(i)}$ represents all rows of $\mathbf{L}^{(i)}$ up to, but not including, the $k$-th one.} and parametrize $p(\mathbf{\hat{L}}_{k}^{(i)}|\mathbf{x}^{(i)}, \mathbf{L}_{<k}^{(i)})$ as a categorical distribution. In the literature, these conditionals are usually modeled with auto-regressive (recurrent) models. In our experiments, we choose the transformer model as well. It is worth mentioning that a potential drawback of this formulation is that it inherently penalizes for order, which might not necessarily be relevant for ingredients.

A \emph{set of ingredients} is a variable sized, unordered collection of unique meal constituents. We can obtain a set of ingredients $S$ by selecting $K$ ingredients from the dictionary $\mathcal{D}$: $S = \{s_i\}_{i=0}^{K}$. We represent $S$ as a binary vector $\mathbf{s}$ of dimension $N$, where $\mathbf{s}_i=1$ if $\mathbf{s}_i \in S$ and $0$ otherwise. Thus, our training data consists of $M$ image and ingredient set pairs: $\{(\mathbf{x}^{(i)}, \mathbf{s}^{(i)})\}_{i=0}^{M}$. In this case, the goal is to predict $\mathbf{\hat{s}}$ from an image $\mathbf{x}$ by maximizing the following objective:
\vspace{-1.5mm}
\begin{equation}
   \argmax_{\theta_I, \theta_L} \sum_{i=0}^M \log p(\mathbf{\hat{s}}^{(i)}=\mathbf{s}^{(i)}|\mathbf{x}^{(i)};\theta_I, \theta_L).
\label{eq:ing_set}
\end{equation}
\vspace{-2mm}

Assuming independence among elements, we can factorize $p(\mathbf{\hat{s}}^{(i)}=\mathbf{s}^{(i)}|\mathbf{x}^{(i)})$ as $\sum_{j=0}^N \log p(\mathbf{\hat{s}}_j^{(i)}=\mathbf{s}_j^{(i)}|\mathbf{x}^{(i)})$. However, the ingredients in the set are not necessarily independent, e.g. \emph{salt} and \emph{pepper} frequently appear together.

\begin{figure}
  \includegraphics[width=0.95\columnwidth]{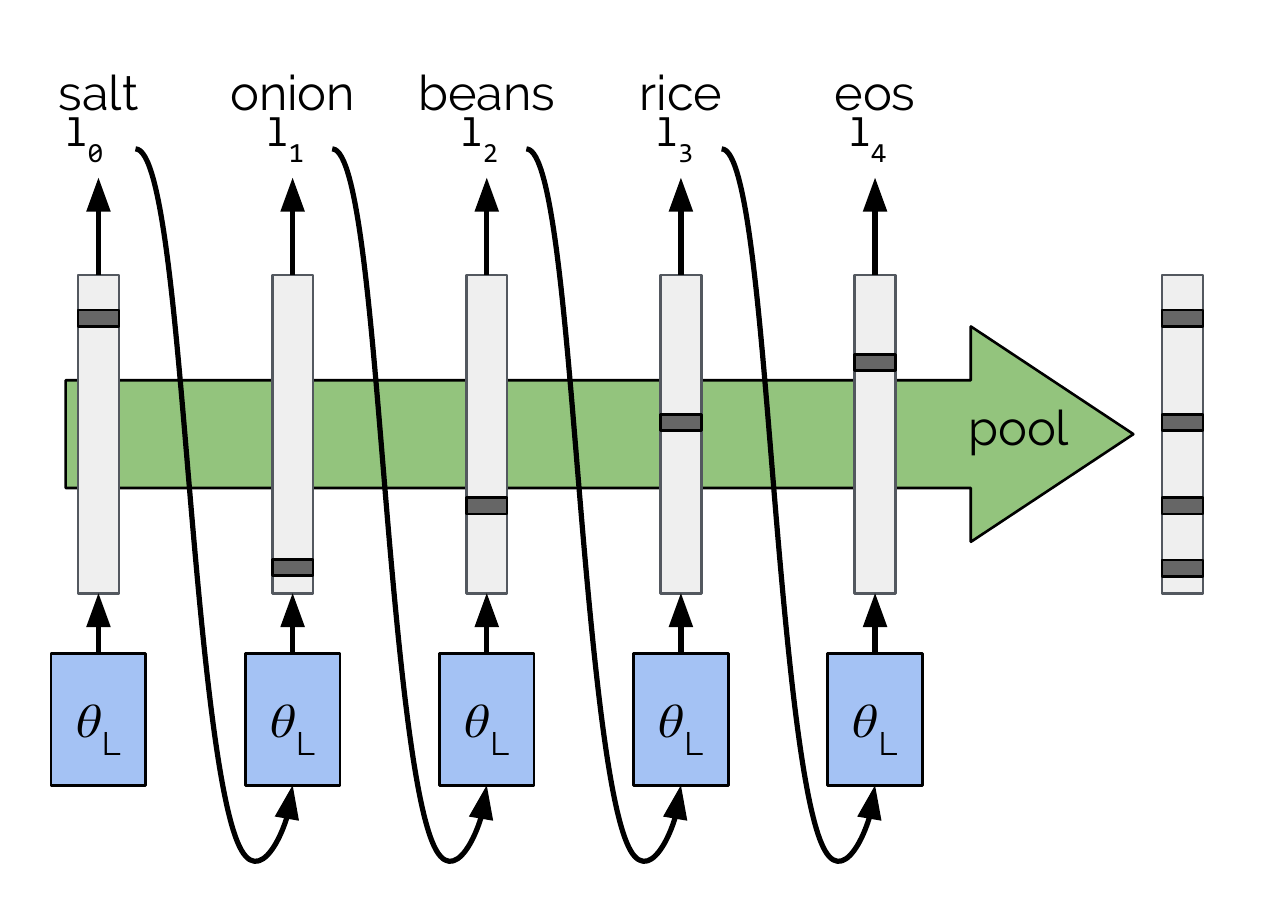}
  \caption{\textbf{Set transformer ($\mathrm{TF}_{set}$).} Softmax probabilities are pooled across time to avoid penalizing for order. }
  \label{fig:hybrid}
  \vspace{-0.3cm}
\end{figure}

To account for element dependencies in the set, we model the set as a list, i.e. as a product of conditional probabilities, by means of an auto-regressive model such as the transformer. The transformer predicts ingredients in a list-like fashion $p(\mathbf{\hat{L}}_{k}^{(i)}|\mathbf{x}^{(i)}, \mathbf{L}_{<k}^{(i)})$, until the end of sequence $eos$ token is encountered. As mentioned previously, the drawback of this approach is that such model design \emph{penalizes for order}. In order to remove the order in which ingredients are predicted, we propose to aggregate the outputs across different time-steps by means of a max pooling operation (see Figure \ref{fig:hybrid}). Moreover, to ensure that the ingredients in $\mathbf{\hat{L}}^{(i)}$ are selected without repetition, we force the pre-activation of $p(\mathbf{\hat{L}}_{k}^{(i)}|\mathbf{x}^{(i)}, \mathbf{L}_{<k}^{(i)})$ to be $- \infty $ for all previously selected ingredients at time-steps $<k$. We train this model by minimizing the binary cross-entropy between the predicted ingredients (after pooling) and the ground truth. Including the $eos$ in the pooling operation would result in loosing the information of where the token appears. Therefore, in order to learn the stopping criteria of the ingredient prediction, we introduce an additional loss accounting for it. The $eos$ loss is defined as the binary cross-entropy loss between the predicted $eos$ probability at all time-steps and the ground truth (represented as a unit step function, whose value is $0$ for the time-steps corresponding to ingredients and $1$ otherwise). In addition to that, we incorporate a cardinality $\ell_1$ penalty, which we found empirically useful. At inference time, we directly sample from the transformer's output. We refer to this model as \emph{set transformer}.

Alternatively, we could use \emph{target distribution} $p(\mathbf{s}^{(i)}|\mathbf{x}^{(i)}) = \mathbf{s}^{(i)}/{\sum_j \mathbf{s}_j^{(i)}}$ \cite{GongJLTI13,Mahajan18hashtags} to model the joint distribution of set elements and train a model by minimizing the cross-entropy loss between $p(\mathbf{s}^{(i)}|\mathbf{x}^{(i)})$ and the model's output distribution $p(\mathbf{\hat{s}}^{(i)}|\mathbf{x}^{(i)})$. Nonetheless, it is not clear how to convert the target distribution back to the corresponding set of elements with variable cardinality. In this case, we build a feed forward network and train it with the target distribution cross-entropy loss. To recover the ingredient set, we propose to greedily sample elements from a \emph{cumulative distribution of sorted output probabilities} $p(\mathbf{\hat{s}}^{(i)}|\mathbf{x}^{(i)})$ and stop the sampling once the sum of probabilities of selected elements is above a threshold. We refer to this model as \emph{feed forward (target distribution)}.

\subsection{Optimization}
\label{sec_optim}
We train our recipe transfomer in two stages. In the first stage, we pre-train the image encoder and ingredients decoder as presented in Subsection \ref{sec_ingrpred}. Then, in the second stage, we train the ingredient encoder and instruction decoder (following Subsection \ref{sec_recipegen}) by minimizing the negative log-likelihood and adjusting $\theta_R$ and $\theta_E$. Note that, while training, the instruction decoder takes as input the ground truth ingredients. All transformer models are trained with teacher forcing \cite{Williams:1989} except for the set transformer.

\section{Experiments}
\label{sec_experiments}
This section is devoted to the dataset and the description of implementation details, followed by an exhaustive analysis of the proposed attention strategies for the cooking instruction transformer. Further, we quantitatively compare the proposed ingredient prediction models to previously introduced baselines. Finally, a comparison of our inverse cooking system with retrieval-based models as well as a comprehensive user study is provided.

\subsection{Dataset}
We train and evaluate our models on the Recipe1M dataset \cite{recipe1m}, composed of $1\,029\,720$ recipes scraped from cooking websites. The dataset contains $720\,639$ training, $155\,036$ validation and $ 154\,045$ test recipes, containing a title, a list of ingredients, a list of cooking instructions and (optionally) an image. In our experiments, we use only the recipes containing images, and remove recipes with less than $2$ ingredients or $2$ instructions, resulting in $252\,547$ training, $54\,255$ validation and $54\,506$ test samples.

Since the dataset was obtained by scraping cooking websites, the resulting recipes are highly unstructured and contain frequently redundant or very narrowly defined cooking ingredients (e.g. \emph{olive oil}, \emph{virgin olive oil} and \emph{spanish olive oil} are separate ingredients). Moreover, the ingredient vocabulary contains more than $400$ different types of \emph{cheese}, and more than $300$ types of \emph{pepper}. As a result, the original dataset contains $16\,823$ unique ingredients, which we preprocess to reduce its size and complexity. First, we merge ingredients if they share the first or last two words (e.g. \emph{bacon cheddar cheese} is merged into \emph{cheddar cheese}); then, we cluster the ingredients that have same word in the first or in the last position (e.g. \emph{gorgonzola cheese} or \emph{cheese blend} are clustered together into the \emph{cheese} category); finally we remove plurals and discard ingredients that appear less than $10$ times in the dataset. Altogether, we reduce the ingredient vocabulary from over $16k$ to $1\,488$ unique ingredients. For the cooking instructions, we tokenize the raw text and remove words that appear less than $10$ times in the dataset, and replace them with unknown word token. Moreover, we add special tokens for the start and the end of recipe as well as the end of instruction. This process results in a recipe vocabulary of $23\,231$ unique words.

\subsection{Implementation Details}
We resize images to $256$ pixels in their shortest side and take random crops of $224\times224$ for training and we select central $224\times224$ pixels for evaluation. For the instruction decoder, we use a transformer with $16$ blocks and $8$ multi-head attentions, each one with dimensionality $64$. For the ingredient decoder, we use a transformer with $4$ blocks and $2$ multi-head attentions, each one with dimensionality of $256$. To obtain image embeddings we use the last convolutional layer of ResNet-50 model. Both image and ingredients embedings are of dimension $512$. We keep a maximum of $20$ ingredients per recipe and truncate instructions to a maximum of $150$ words. The models are trained with Adam optimizer \cite{KingmaB14} until early-stopping criteria is met (using patience of $50$ and monitoring validation loss). All models are implemented with PyTorch\footnote{\url{https://pytorch.org/}}~\cite{paszke2017automatic}. Additional implementation details are provided in the supplementary material.

\subsection{Recipe Generation}
\begin{table}
\small
\centering
\begin{tabular}[b]{lc}
\toprule
Model            & ppl \\ \midrule
Independent      &    $8.59$          \\
Seq. img. first  &    $8.53$          \\
Seq. ing. first  &    $8.61$          \\ 
Concatenated     &    $\mathbf{8.50}$  \\ \bottomrule
\end{tabular}
\quad
\centering
\begin{tabular}[b]{lcc}
    \toprule
                Model & IoU & $\mathrm{F1}$ \\ \midrule
            $\mathrm{FF}_{BCE}$                 & 17.85 & 30.30 \\
            $\mathrm{FF}_{IOU}$                & 26.25 & 41.58 \\
            $\mathrm{FF}_{DC}$                  & 27.22 & 42.80  \\
            $\mathrm{FF}_{TD}$   & \textbf{28.84} & \textbf{44.11} \\\midrule
            $\mathrm{TF}_{list}$                 & 29.48 & 45.55 \\
            $\mathrm{TF}_{list}$ + shuf.       & 27.86 & 43.58 \\
            $\mathrm{TF}_{set}$  & \textbf{31.80} & \textbf{48.26} \\ \bottomrule
    \end{tabular}
    
\caption{\textbf{Model selection (val).} \underline{Left}: Recipe perplexity (ppl). \underline{Right}: Global ingredient IoU \& F1. }
\label{tab:model_selection}
\vspace{-0.3cm}
\end{table}

In this section, we compare the proposed multi-modal attention architectures described in Section \ref{sec_recipegen}. Table \ref{tab:model_selection} (left) reports the results in terms of perplexity on the validation set. We observe that independent attention exhibits the lowest results, followed by both sequential attentions. While the latter have the capability to refine the output with either ingredient or image information consecutively, independent attention can only do it in one step. This is also the case of concatenated attention, which achieves the best performance. However, concatenated attention is flexible enough to decide whether to give more focus to one modality, at the expense of the other, whereas independent attention is forced to include information from both modalities.
Therefore, we use the concatenated attention model to report results on the test set. We compare it to a system going directly from image-to-sequence of instructions without predicting ingredients (I2R). Moreover, to assess the influence of visual features on recipe quality, we adapt our model by removing visual features and predicting instructions directly from ingredients (L2R). Our system achieves a test set perplexity of $8.51$, improving both I2R and L2R baselines, and highlighting the benefits of using both image and ingredients when generating recipes. L2R surpasses I2R with a perplexity of $8.67$ vs. $9.66$, demonstrating the usefulness of having access to concepts (ingredients) that are essential to the cooking instructions. Finally, we greedily sample instructions from our model and analyze the results. We notice that generated instructions have an average of $9.21$ sentences containing $9$ words each, whereas real, ground truth instructions have an average of $9.08$ sentences of length $12.79$. See supplementary material for qualitative examples of generated recipes.

\subsection{Ingredient Prediction}


\begin{figure*}[t]
    \begin{minipage}{0.33\linewidth}
    \small
    \centering
    \resizebox{\columnwidth}{!}{

    \begin{tabular}[b]{@{}lcc@{}}
    \toprule
      &  Card. error &  \# pred. ingrs \\
      \midrule
     $\mathrm{FF}_{BCE}$ &  $5.67 \pm{3.10}$ &  $2.37 \pm{1.58}$ \\
     $\mathrm{FF}_{DC}$  &  $2.68 \pm{2.07}$ &  $9.18 \pm{2.06}$ \\
     $\mathrm{FF}_{IOU}$   & $2.46 \pm{1.95}$ &   $7.86 \pm{1.72}$ \\
     $\mathrm{FF}_{TD}$  &  $3.02 \pm{2.50}$ &  $8.02 \pm{3.24}$ \\
     \midrule
     $\mathrm{TF}_{list}$  & $2.49 \pm{2.11}$ &  $7.05 \pm{2.77}$ \\
     $\mathrm{TF}_{list}$ + shuffle &  $3.24 \pm{2.50}$  &  $5.06 \pm{1.85}$ \\
     $\mathrm{TF}_{set}$  &  $2.56 \pm{1.93}$ &  $9.43 \pm{2.35}$ 
 \\\bottomrule
    \end{tabular}
    }
    \vspace{-1mm}
        \captionof{table}{\textbf{Ingredient Cardinality.}}
        \label{tab:ingr_cardinality}
    \end{minipage}
    \begin{minipage}{0.66\linewidth}
      \begin{subfigure}[b]{0.5\textwidth}
        \includegraphics[width=\textwidth]{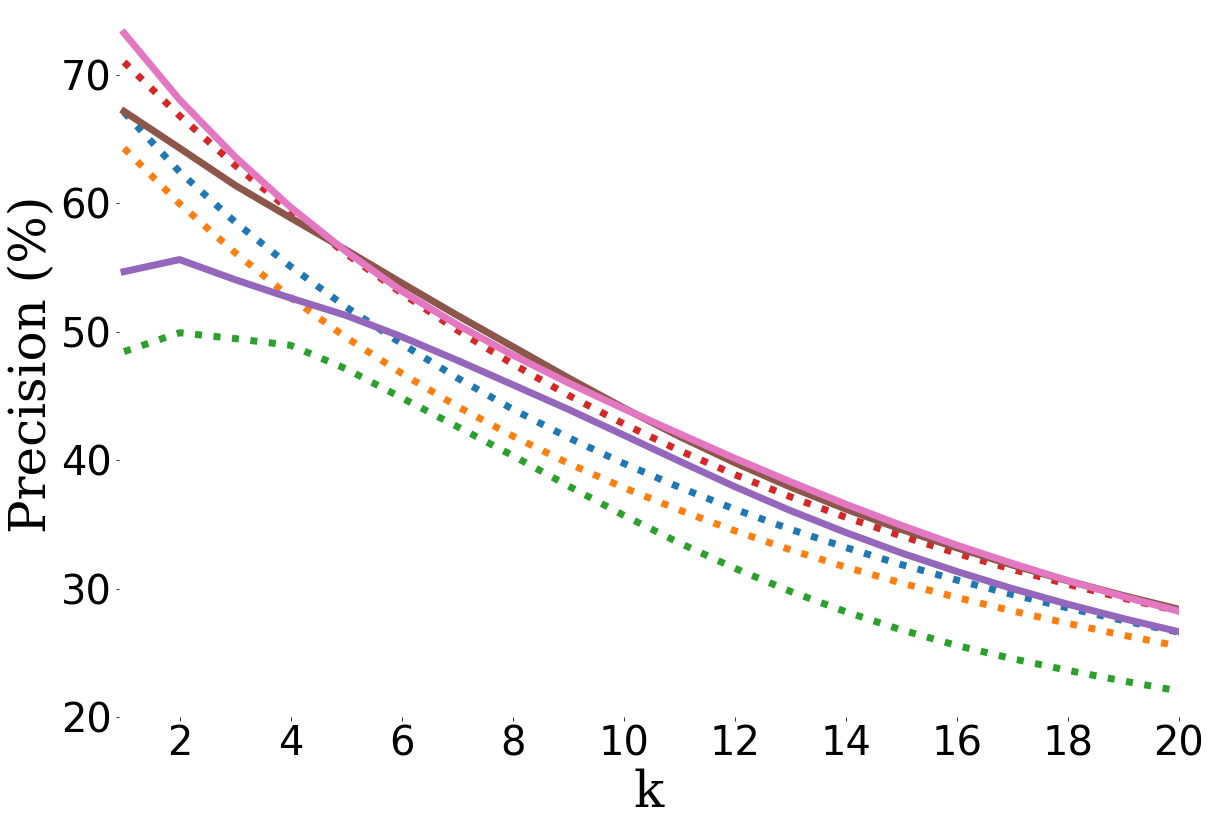}
        \vspace{-.7cm}
        \label{fig:f1}
    \end{subfigure}
    ~ 
    \begin{subfigure}[b]{0.5\textwidth}
        \includegraphics[width=\textwidth]{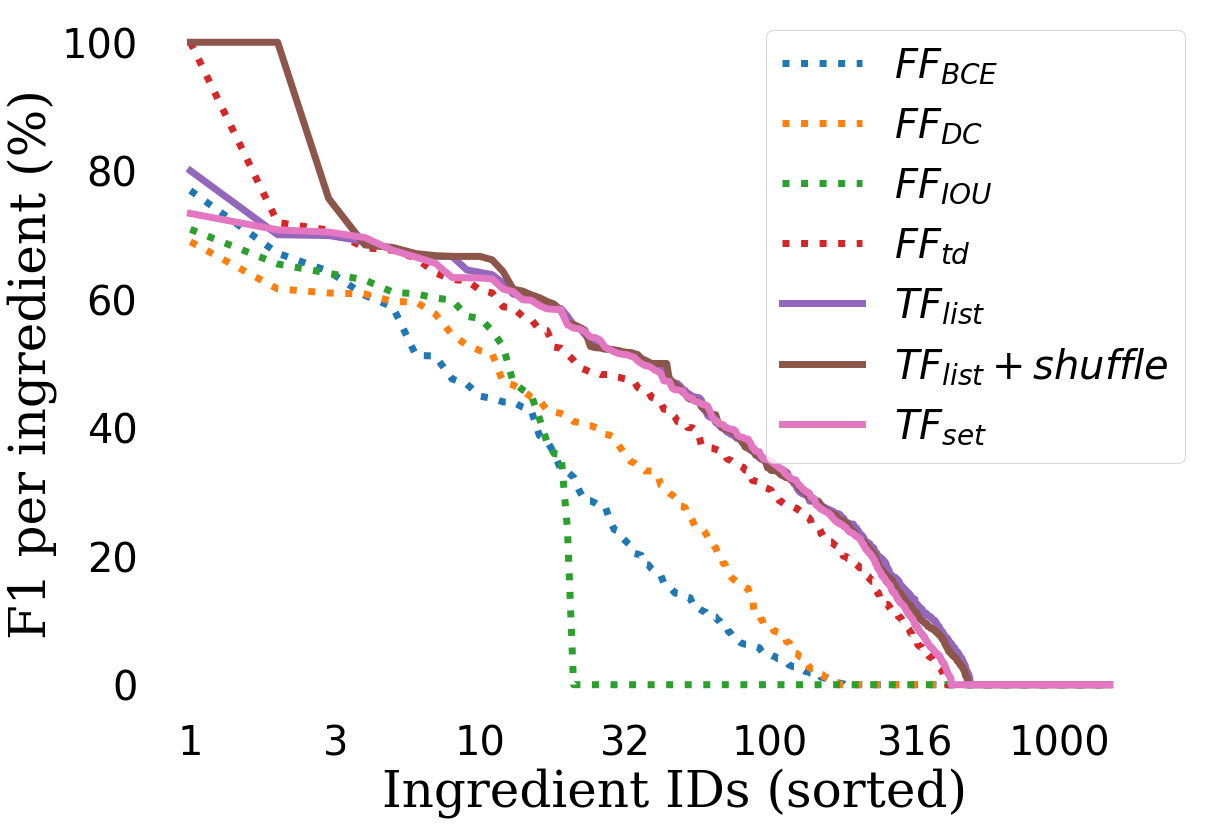}
        \vspace{-.7cm}
        \label{fig:f1ingr}
    \end{subfigure}
    \caption{\textbf{Ingredient prediction results:} P@K and F1 per ingredient.}\label{fig:exp_plots}    
    \end{minipage} 
    \vspace{-.1cm}
\end{figure*}
In this section, we compare the proposed ingredient prediction approaches to previously introduced models, with the goal of assessing whether ingredients should be treated as lists or sets. We consider models from the multilabel classification literature as baselines, and tune them for our purposes. On the one hand, we have models based on feed forward convolutional networks, which are trained to predict sets of ingredients. We experiment with several losses to train these models, namely binary cross-entropy, soft intersection over union as well as target distribution cross-entropy. Note that binary cross-entropy is the only one not taking into account dependencies among elements in the set. On the other hand, we have sequential models that predict lists, imposing order and exploiting dependencies among elements. Finally, we consider recently proposed models which couple set prediction with cardinality prediction to determine which elements to include in the set \cite{rezatofighi2017joint}.

Table \ref{tab:model_selection} (right) reports the results on the validation set for the state-of-the-art baselines as well as the proposed approaches. We evaluate the models in terms of Intersection over Union (IoU) and F1 score, computed for accumulated counts of $TP$, $FN$ and $FP$ over the entire dataset split (following Pascal VOC convention). As shown in the table, the feed forward model trained with binary cross-entropy \cite{chen2016deep} ($\mathrm{FF}_{BCE}$) exhibits the lowest performance on both metrics, which could be explained by the assumed independence among ingredients. These results are already notably improved by the method that learns to predict the set cardinality ($\mathrm{FF}_{DC}$). Similarly, the performance increases when training the model with structured losses such as soft IoU ($\mathrm{FF}_{IOU}$). Our feed forward model trained with target distribution ($\mathrm{FF}_{TD}$) and sampled by thresholding (th = $0.5$) the sum of probabilities of selected ingredients outperforms all feed forward baselines, including recently proposed alternatives for set prediction such as \cite{rezatofighi2017joint} ($\mathrm{FF}_{DC}$). Note that target distribution models dependencies among elements in a set and implicitly captures cardinality information. Following recent literature modeling sets as lists \cite{Nam17setrnn}, we train a transformer network to predict ingredients given an image by minimizing the negative log-likelihood loss ($\mathrm{TF}_{list}$). Moreover, we train the same transformer by randomly shuffling the ingredients (thus, removing order from the data). Both models exhibit competitive results when compared to feed forward models, highlighting the importance of modeling dependencies among ingredients. Finally, our proposed set transformer $\mathrm{TF}_{set}$, which models ingredient co-occurrences exploiting the auto-regressive nature of the model yet satisfying order invariance, achieves the best results, emphasizing the importance of modeling dependencies, while not penalizing for any given order.

The average number of ingredients per sample in Recipe1M is \emph{$7.99 \pm{3.21}$} after pre-processing. We report the cardinality prediction errors as well as the average number of predicted ingredients for each of the tested models in Table \ref{tab:ingr_cardinality}. $\mathrm{TF}_{set}$ is the third best method in terms of cardinality error (after $\mathrm{FF}_{IOU}$ and $\mathrm{TF}_{list}$), while being superior to all methods in terms of F1 and IoU. Further, Figure \ref{fig:exp_plots} (left) shows the precision score at different values of $K$. As observed, the plot follows similar trends as Table \ref{tab:model_selection} (right), with $\mathrm{FF}_{TD}$ being among the most competitive models and $\mathrm{TF}_{set}$ outperforming all previous baselines for most values of $K$. Figure \ref{fig:exp_plots} (right) shows the F1 per ingredient, where the ingredients in the horizontal axes are sorted by score. Again, we see that models that exploit dependencies consistently improve ingredient's F1 scores, strengthening the importance of modeling ingredient co-occurrences. 


\subsection{Generation vs Retrieval}

In this section, we compare our proposed recipe generation system with retrieval baselines, which we use to search recipes in the \emph{entire} test set for fair comparison.

\begin{figure}
  \includegraphics[width=\columnwidth]{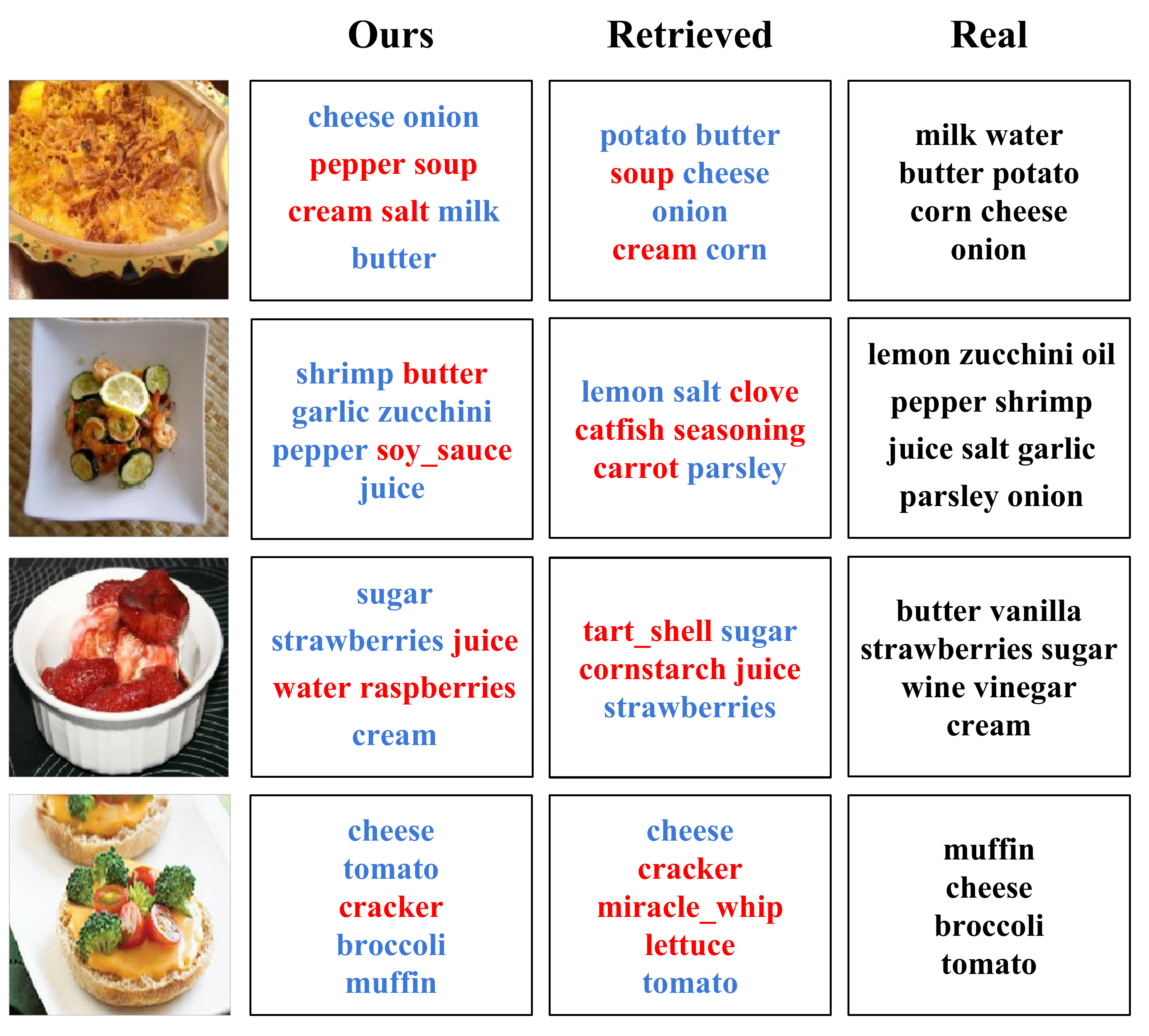}
  \caption{\textbf{Ingredient prediction examples. } We compare obtained ingredients with our method and the retrieval baseline. Ingredients are displayed in blue if they are present in the real sample and red otherwise. Best viewed in color.}
  \label{fig:ingredient_pred}
   \vspace{-3mm}
\end{figure}

\textbf{Ingredient prediction evaluation.} We use the retrieval model in \cite{recipe1m} as a baseline and compare it with our best ingredient predictions models, namely $\mathrm{FF}_{TD}$ and $\mathrm{FF}_{set}$. The retrieval model, which we refer to as $\mathrm{R}_{I2LR}$, learns joint embeddings of images and recipes (title, ingredients and instructions). Therefore, for the ingredient prediction task, we use the image embeddings to retrieve the closest recipe and report metrics for the ingredients of the retrieved recipe. We further consider an alternative retrieval architecture, which learns joint embeddings between images and ingredients list (ignoring title and instructions). We refer to this model as $\mathrm{R}_{I2L}$. Table \ref{tab:ingr_test} (left) reports the obtained results on the Recipe1M test set. The $\mathrm{R}_{I2LR}$ model outperforms the $\mathrm{R}_{I2L}$ one, which indicates that instructions contain complementary information that is useful when learning effective embeddings. Furthermore, both of our proposed methods outperform the retrieval-baselines by a large margin (e.g. $\mathrm{TF}_{set}$ outperforms the $\mathrm{R}_{I2LR}$ retrieval baseline by $12.26$ IoU points and $15.48$ F1 score points), which demonstrates the superiority of our models. Finally, Figure \ref{fig:ingredient_pred} presents some qualitative results for image-to-ingredient prediction for our model as well as for the retrieval based system. We use blue to highlight the ingredients that are present in the ground truth annotation and red otherwise. 

\begin{table}[]
\small
\centering
\begin{tabular}[b]{lcc}
\toprule
                 & IoU & $\mathrm{F1}$ \\ \midrule
    $\mathrm{R}_{I2L}$ \cite{recipe1m}  & 18.92    &  31.83  \\
    $\mathrm{R}_{I2LR}$ \cite{recipe1m} & 19.85    &  33.13   \\ \midrule
    $\mathrm{FF}_{TD}$ (ours)        & 29.82    &  45.94 \\
    $\mathrm{TF}_{set}$ (ours)      & \textbf{32.11}    & \textbf{48.61}  \\ \bottomrule
\end{tabular}
\quad
\centering
\begin{tabular}[b]{@{}lcc@{}}
\toprule
            & Rec. & Prec. \\ \midrule
$\mathrm{R}_{\mathrm{IL2R}}$  &     $31.92$   & $28.94$        \\
Ours & $\mathbf{75.47}$     &   $\mathbf{77.13}$    \\ \bottomrule
 
\end{tabular}
\caption{\textbf{Test performance against retrieval.} Left: Global ingredient IoU and F1 scores. Right: Precision and Recall of ingredients in cooking instructions.}
\label{tab:ingr_test}
 \vspace{-0.3cm}
\end{table}

\begin{table}
\small
\centering
\begin{tabular}[b]{@{}ccc@{}}
\toprule
    & IoU &  F1  \\ \midrule
Human &  21.36    & 35.20  \\
Retrieved  &   18.03   &  30.55  \\ 
Ours &  \textbf{32.52}    & \textbf{49.08} \\ \bottomrule
\end{tabular}
\quad
\begin{tabular}[b]{@{}lc@{}}
\toprule
           & Success \% \\ \midrule
Real            &         80.33       \\
Retrieved &    48.81           \\
Ours            &         55.47       \\ \bottomrule 
\end{tabular}
\caption{\textbf{User studies.} Left: IoU \& F1 scores for ingredients obtained with retrieval \cite{recipe1m}, our approach and humans. Right: Recipe success rate according to human judgment.}
\vspace{-0.2cm}
\label{tab:human}
\end{table}

\textbf{Recipe generation evaluation.} We compare our proposed instruction decoder (which generates instructions given an image and ingredients) with a retrieval variant. For a fair comparison, we retrain the retrieval system to find the cooking instructions given both image and ingredients. In our evaluation, we consider the ground truth ingredients as reference and compute \emph{recall} and \emph{precision} w.r.t. the ingredients that appear in the obtained instructions. Thus, recall computes the percentage of ingredients in the reference that appear in the output instructions, whereas precision measures the percentage of ingredients appearing in the instructions that also appear in the reference. Table \ref{tab:ingr_test} (right) displays comparison between our model and the retrieval system. Results show that ingredients appearing in generated instructions have better recall and precision scores than the ingredients in retrieved instructions. 

\subsection{User Studies}
\label{sec:experiments_users}
In this section, we quantify the quality of predicted ingredients and generated instructions with user studies. In the first study, we compare the performance of our model against human performance in the task of recipe generation (including ingredients and recipe instructions). We randomly select $15$ images from the test set, and ask users to select up to $20$ distinct ingredients as well as write a recipe that would correspond with the provided image. To reduce the complexity of the task for humans, we reduced the ingredient vocabulary from $1\,488$ to $323$, by increasing the frequency threshold from $10$ to $1$k. We collected answers from $31$ different users, altogether collecting an average of $5.5$ answers for each image. For fair comparison, we re-train our best ingredient prediction model on the reduced vocabulary of ingredients. We compute IoU and F1 ingredient scores obtained by humans, the retrieval baseline and our method. Results are included in Table \ref{tab:human} (left), underlining the complexity of the task. As shown in the table, humans outperform the retrieval baseline (F1 of $35.20\%$ vs $30.55\%$, respectively). Furthermore, our method outperforms both human baseline and retrieval based systems obtaining F1 of $49.08\%$. Qualitative comparisons between generated and human-written recipes (including recipes from average and expert users) are provided in the supplementary material.

The second study aims at quantifying the quality of the generated recipes (ingredients and instructions) with respect to (1) the real recipes in the dataset, and (2) the ones obtained with the retrieval baseline \cite{recipe1m}. With this purpose, we randomly select $150$ recipes with their associated images from the test set and, for each image, we collect the corresponding real recipe, the top-1 retrieved recipe and our generated recipe. We present the users with $15$ image-recipe pairs (randomly chosen among the real, retrieved and generated ones) asking them to indicate whether the recipe matches the image. In the study, we collected answers from 105 different users, resulting in an average of $10$ responses for each image. Table \ref{tab:human} (right) presents the results of this study, reporting the success rate of each recipe type. As it can be observed, the success rate of generated recipes is higher than the success rate of retrieved recipes, stressing the benefits of our approach w.r.t. retrieval.

\section{Conclusion}
\label{sec_conclusion}

In this paper, we introduced an image-to-recipe generation system, which takes a food image and produces a recipe consisting of a title, ingredients and sequence of cooking instructions. We first predicted sets of ingredients from food images, showing that modeling dependencies matters. Then, we explored instruction generation conditioned on images and inferred ingredients, highlighting the importance of reasoning about both modalities at the same time. Finally, user study results confirm the difficulty of the task, and demonstrate the superiority of our system against state-of-the-art image-to-recipe retrieval approaches.

\section{Acknowledgements}
\label{sec_acks}

We are grateful to Nicolas Ballas, Lluis Castrejon, Zizhao Zhang and Pascal Vincent for their fruitful comments and suggestions. We also want to express our gratitude to Joelle Pineau for her unwavering support to this project. Finally, we wish to thank everyone who anonymously participated in the user studies.

This work has been partially developed in the framework of projects TEC2013-43935-R and TEC2016-75976-R, financed by the Spanish Ministerio de Economía y Competitividad and the European Regional Development Fund.

{\small
\bibliographystyle{ieee}
\bibliography{egbib}
}
\clearpage
\section{Supplementary Material}
\label{sec_supp}

This supplementary material intends to provide further details as well as qualitative results. In Section \ref{sec:sm_impl}, we describe additional implementation and training details. Section \ref{sec:sm_ingrs} presents an analysis of our ingredient vocabulary before and after its pre-processing. Examples of generated recipes, displayed together with real ones from the dataset, are presented in Section \ref{sec:sm_recipes}. Section \ref{sec:sm_forms} includes screenshots of the two forms that were used to collect data for the user studies. Section \ref{sec:sm_written} includes examples of human written recipes compared to real and generated ones. Finally, in Section \ref{sec:sm_dineout}, we provide examples of generated recipes for out-of-dataset pictures taken by authors.

\subsection{Training Details}
\label{sec:sm_impl} 

\textbf{Ingredient Prediction.} Feed-forward models $\mathrm{FF}_{BCE}$, $\mathrm{FF}_{TD}$ and $\mathrm{FF}_{IOU}$ were trained with a mini-batch size of $300$, whereas $\mathrm{FF}_{DC}$ was trained with a mini-batch size of $256$. All of them were trained with a learning rate of $0.001$. The learning rate for pre-trained ResNet layers was scaled for each model as follows: $0.01\times$ for $\mathrm{FF}_{BCE}$, $\mathrm{FF}_{IOU}$ and $\mathrm{FF}_{DC}$ and $0.1\times$ for $\mathrm{FF}_{TD}$. Transformer list-based models $\mathrm{TF}_{list}$ were trained with mini-batch size $300$ and learning rate $0.001$, scaling the learning rate of ResNet layers with a factor of $0.1\times$. Similarly, the set transformer $\mathrm{TF}_{set}$ was trained with mini-batch size of $300$ and a learning rate of $0.0001$, scaling the learning rate of pre-trained ResNet layers with a factor of $1.0\times$. The optimization of $\mathrm{TF}_{set}$ minimizes a cost function composed of three terms, namely the ingredient prediction loss $\mathcal{L}_{ingr}$ and the end-of-sequence loss $\mathcal{L}_{eos}$ and the cardinality penalty $\mathcal{L}_{card}$. We set the contribution of each term with weights $1000.0$ and $1.0$ and $1.0$, respectively. We use a label smoothing factor of $0.1$ for all models trained with BCE loss ($\mathrm{FF}_{BCE}$, $\mathrm{FF}_{DC}$, $\mathrm{TF}_{set}$), which we found experimentally useful. 

\textbf{Instruction Generation.} We use a batch size of $256$ and learning rate of $0.001$. Parameters of the image encoder module are taken from the ingredient prediction model and frozen during training for instruction generation.

All models are trained with Adam optimizer ($\beta_1 = 0.9$, $\beta_1 = 0.99$ and $\epsilon = $1e-8), exponential decay of $0.99$ after each epoch, dropout probability $0.3$ and a maximum number of 400 epochs (if early stopping criterion is not met). During training we randomly flip ($p =0.5$), rotate ($\pm 10$ degrees) and translate images ($\pm 10\%$ image size on each axis) for augmentation.

\subsection{Ingredient Analysis}
\label{sec:sm_ingrs} 

\begin{figure*}
\centering
\caption{\textbf{Ingredient word clouds.} The size of each ingredient word is proportional to the frequency of appearance in the dataset. We display word clouds for ingredients before (\ref{ffig:ingrs_before}) and after (\ref{ffig:ingrs_after}) our pre-processing step. In \ref{ffig:ingrs_cheeses} we show the different types of \emph{cheese} that are clustered together after pre-processing.}
\begin{subfigure}[b]{\textwidth}
    \centering
    \includegraphics[width=0.7\textwidth]{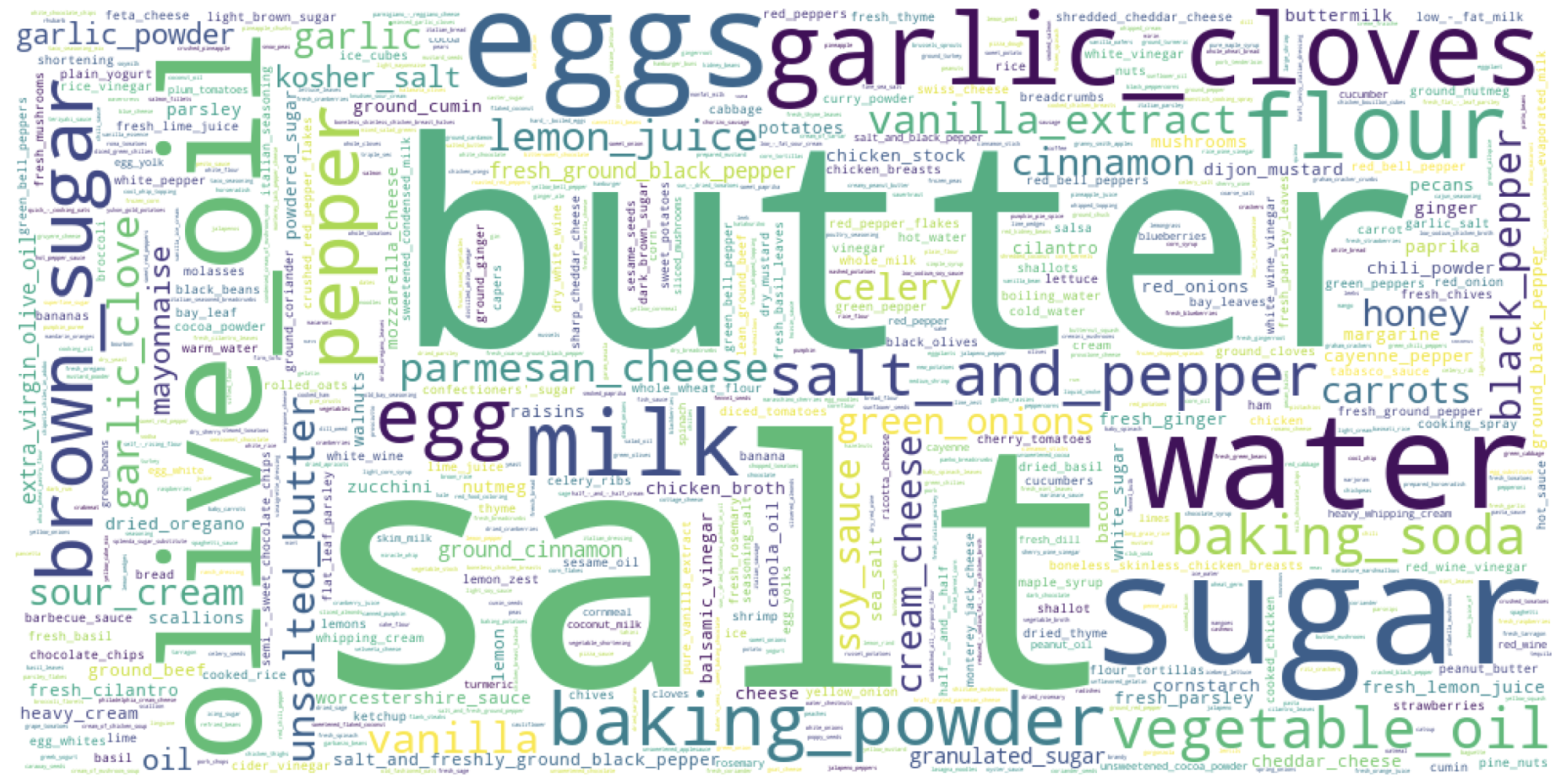}
    \caption{Before pre-processing.}
    \label{ffig:ingrs_before}
\end{subfigure}
\begin{subfigure}[b]{\textwidth}
\centering
    \includegraphics[width=0.7\textwidth]{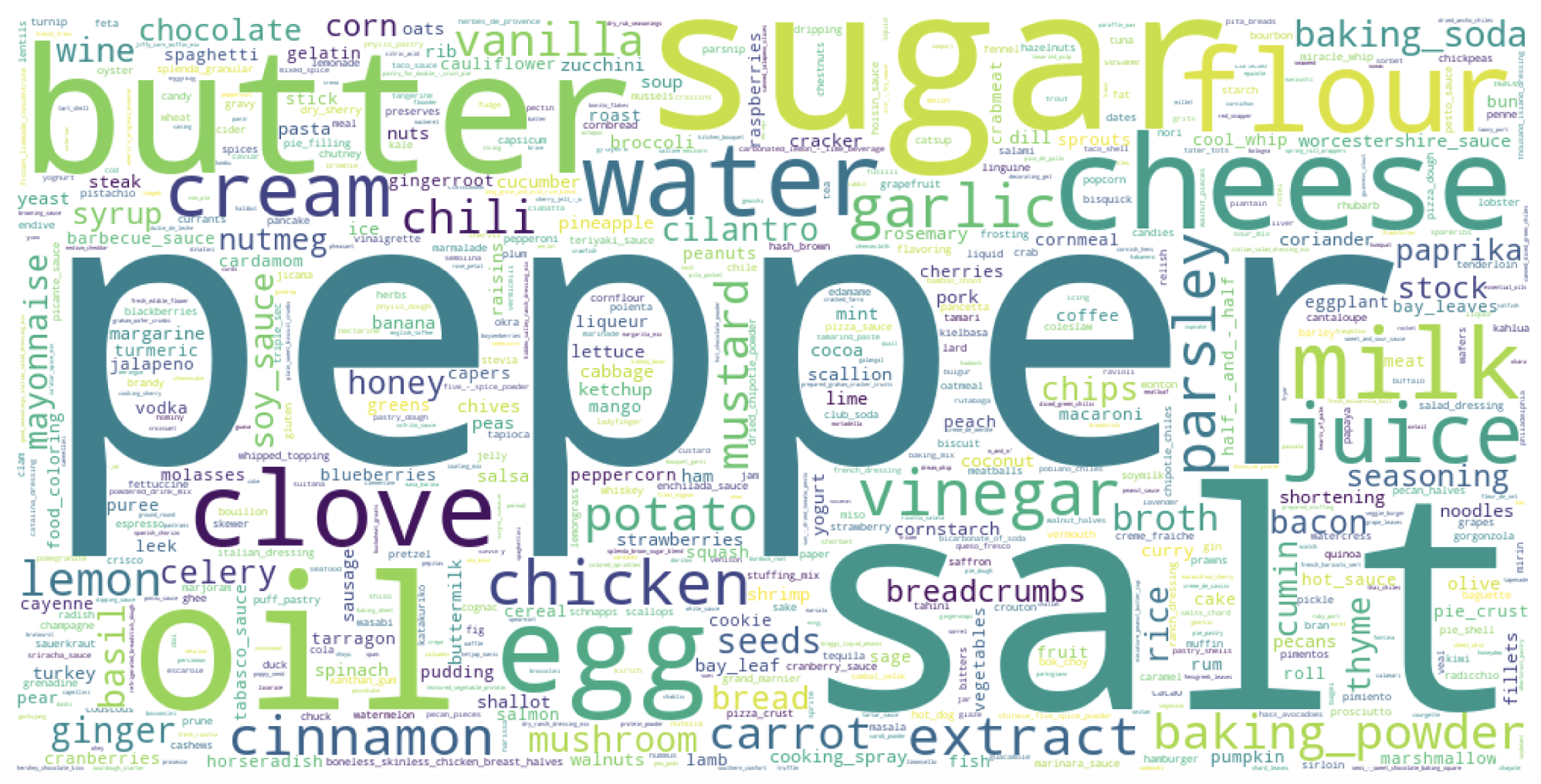}
    \caption{After pre-processing.}
    \label{ffig:ingrs_after}
\end{subfigure}
\begin{subfigure}[b]{\textwidth}
\centering
    \includegraphics[width=0.7\textwidth]{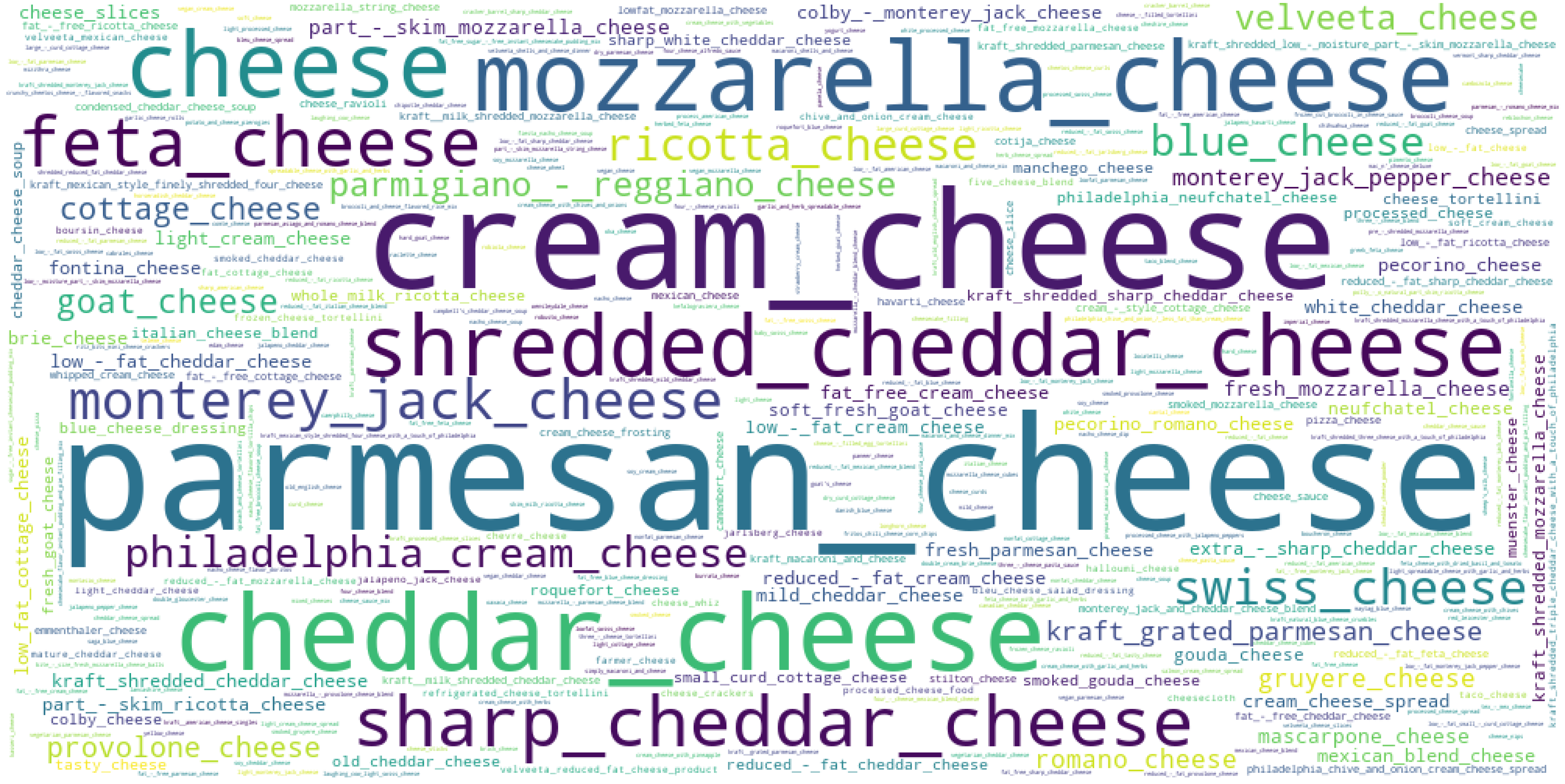}
    \caption{Types of \emph{cheese} before pre-processing.}
    \label{ffig:ingrs_cheeses}
\end{subfigure}
\label{fig:sm_ingrs}
\end{figure*}

We provide visualizations of the ingredient vocabulary used to train our models. Figure \ref{fig:sm_ingrs} displays each unique ingredient in the vocabulary before and after our pre-processing stage. The size of each ingredient word indicates its frequency in the dataset (e.g. \emph{butter} and \emph{salt} appear in many recipes). After filtering and clustering ingredients, the distribution slightly changes (e.g. \emph{pepper} becomes the most frequent ingredient, and popular ingredients such as \emph{olive oil} or \emph{vegetable oil} are clustered into \emph{oil}). Additionally, we illustrate the high ingredient overlap in the dataset with an example of the different types of \emph{cheese} that appear as different ingredients before pre-processing.

\subsection{Generated Recipes}
\label{sec:sm_recipes} 

Figure \ref{fig:sm_recipes} shows additional examples of generated recipes obtained with our method. We also provide the real recipe for completeness. Although sometimes far from the real recipe, our system is able to generate plausible and structured recipes for the input images. Common mistakes include failures in ingredient recognition (e.g. \emph{stuffed tomatoes} are confused with \emph{stuffed peppers} in Figure \ref{fig:recipe_2}), inconsistencies between ingredients and instructions (e.g. \emph{cucumber} is predicted as an ingredient but unused in Figure \ref{fig:recipe_5}, and \emph{meat} is mentioned in the title and instructions but is not predicted as an ingredient in Figure \ref{fig:recipe_8}), and repetitions in ingredient enumeration (e.g. \emph{Stir in tomato sauce, tomato paste, tomato paste, ...} in Figure \ref{fig:recipe_3}).

\subsection{User Study Forms}
\label{sec:sm_forms} 

\begin{figure*}
\centering
  \caption{\textbf{User Study 2.} Recipe quality assessment form.}
  \includegraphics[width=0.9\textwidth ,scale=0.5]{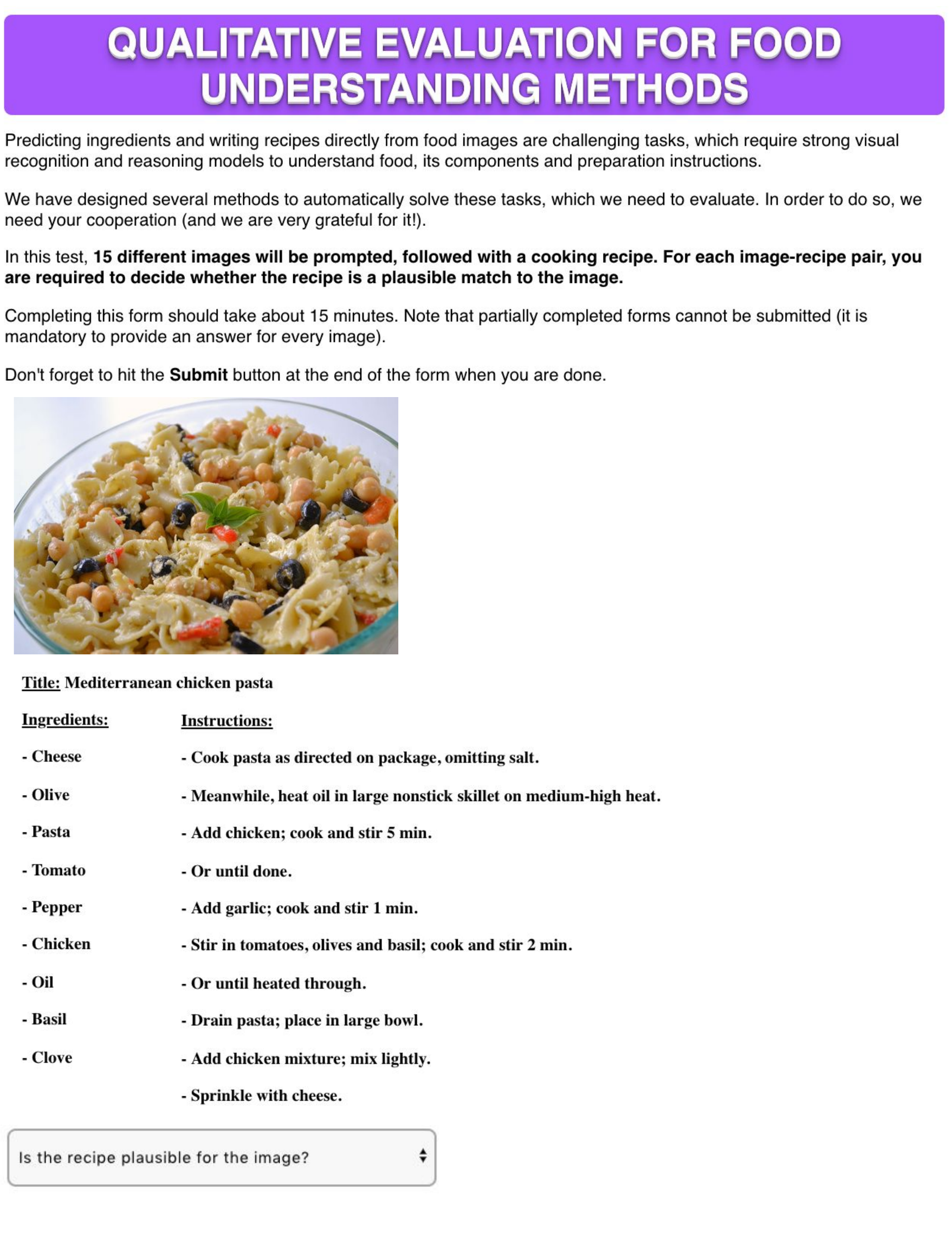}
  \label{fig:study2}
\end{figure*}

We provide screenshots of the two forms used to collect data for user studies. Figure \ref{fig:study1} shows the interface used by users to select image ingredients (each ingredient was selected using a drop-down menu), and write recipes (as free-form text). Figure \ref{fig:study2} shows the form we used to assess whether a recipe matched the provided image according to human judgment.

\subsection{Human-written Recipes}
\label{sec:sm_written} 

\begin{figure*}
\centering
  \caption{\textbf{User Study 1.} Interface for writing recipes and selecting ingredients.}
  \includegraphics[width=0.85\textwidth,scale=0.5]{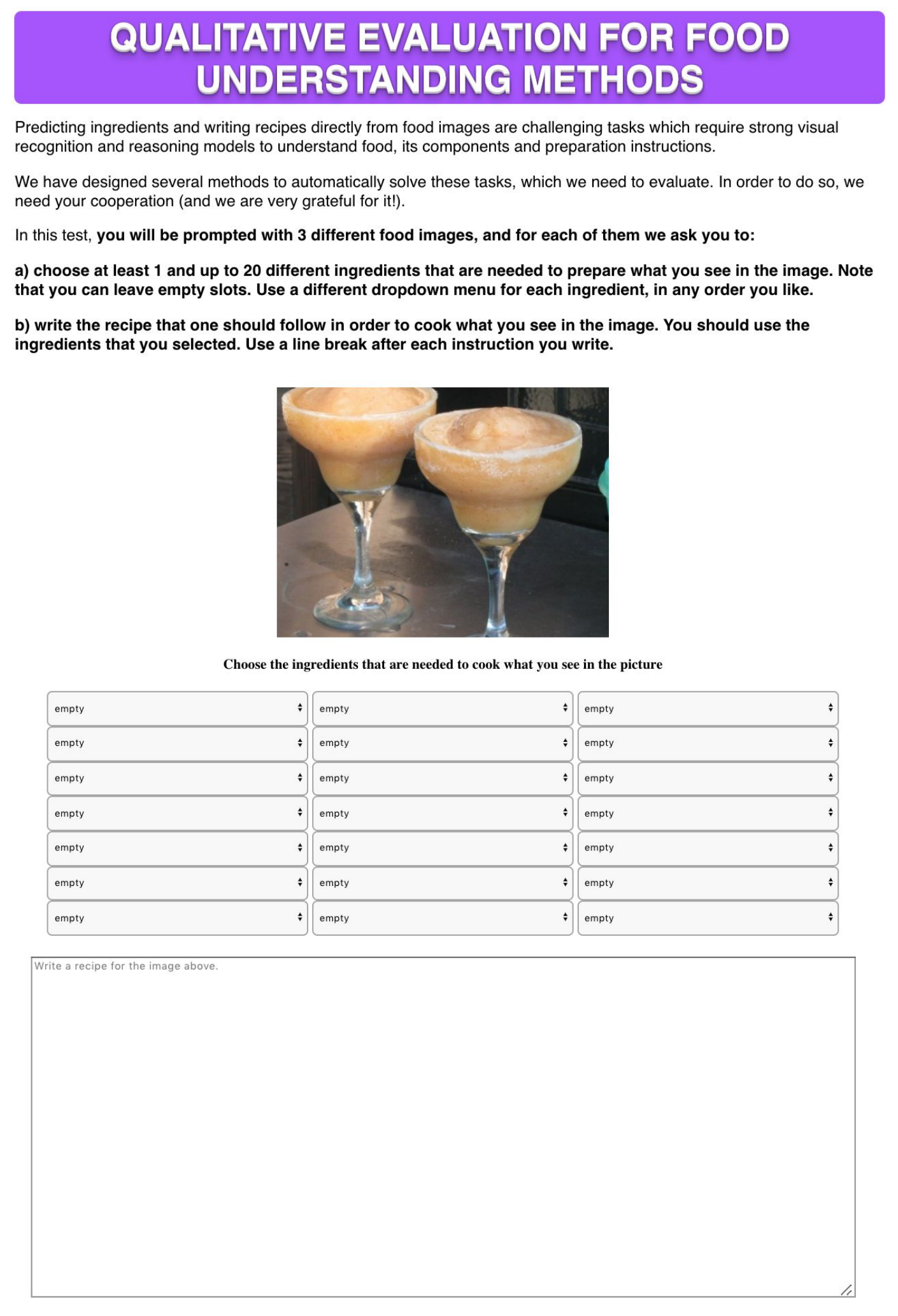}
  \label{fig:study1}
\end{figure*}

\begin{figure*}
\caption{\textbf{Written Recipes.} Real, generated and human written recipes collected with our user study.}
  \includegraphics[width=\textwidth,scale=0.7]{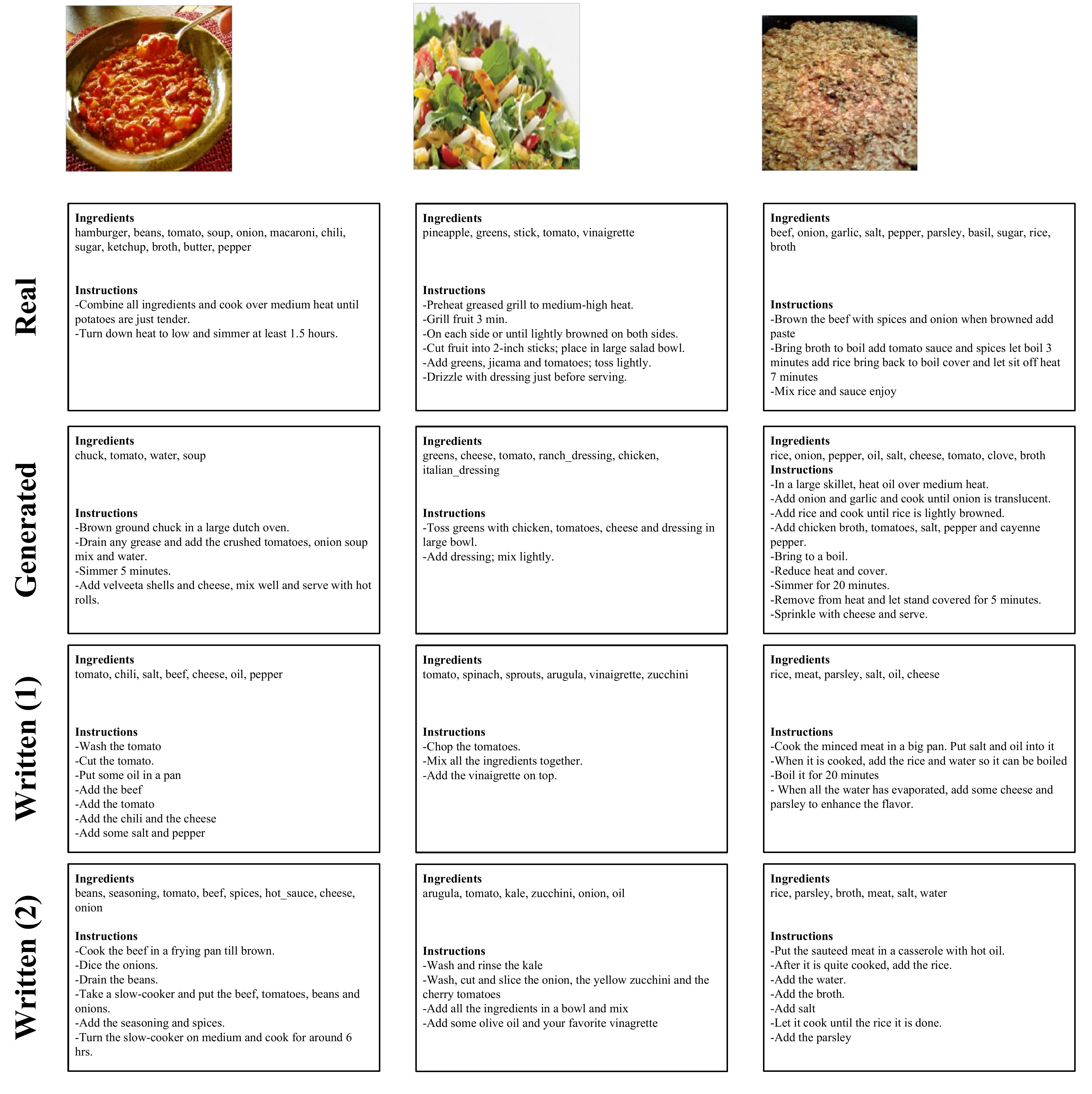}
  \label{fig:written}
\end{figure*}

In Figure \ref{fig:written} we show examples of recipes written by humans, which were collected using the form in Figure \ref{fig:study1}. We also display the real and generated recipes for completeness. Recipes written by humans tend to be shorter, with an average of $5.29$ instructions of $9.03$ words each. In contrast, our model generates recipes that contain an average of instructions $9.21$ of $9$ words each, which closely matches the real distribution ($9.08$ sentences of length $12.79$).

\begin{figure*}
\centering
\caption{\textbf{Recipe examples.} We show both real and generated recipes for different test images.}
\label{fig:sm_recipes}
\begin{subfigure}[b]{\textwidth}
\centering
\includegraphics[trim={0 250 0 0},clip,width=\textwidth]{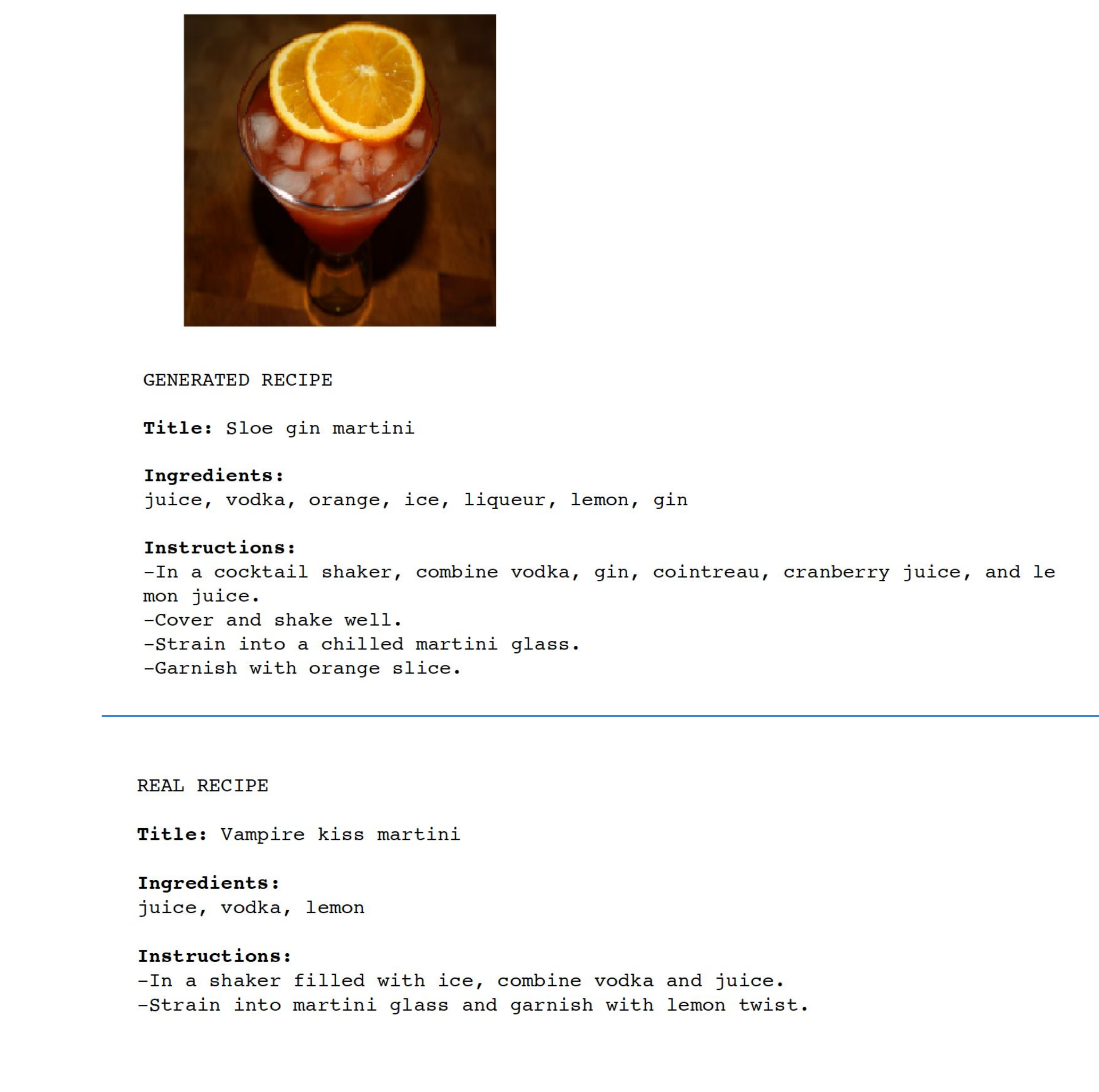}
\caption{\label{fig:recipe_1}}
\end{subfigure}
\end{figure*}

\begin{figure*}
\centering
\ContinuedFloat 
\begin{subfigure}[b]{\textwidth}
\centering
\includegraphics[width=0.9\textwidth]{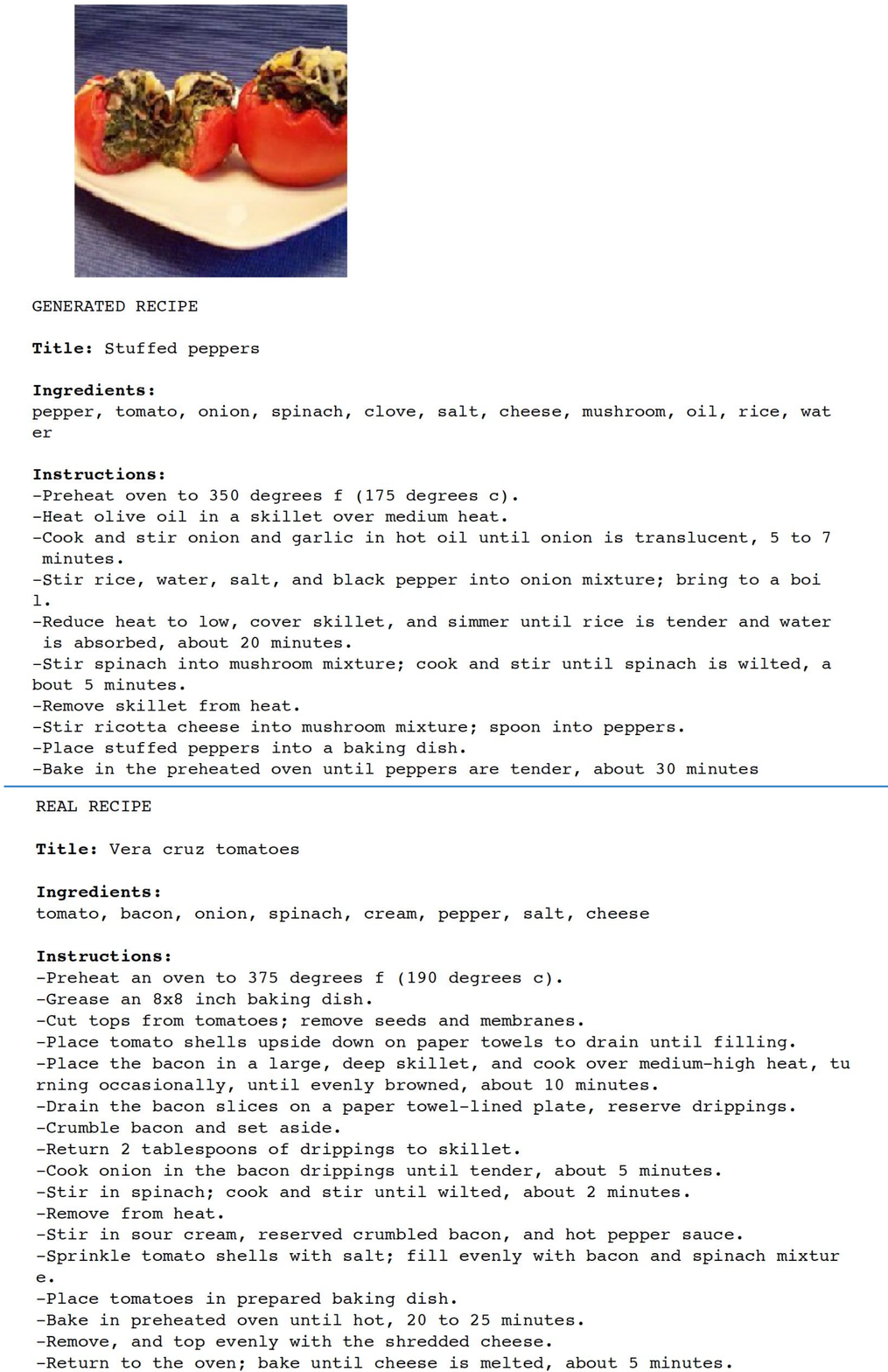}
\caption{\label{fig:recipe_2}}
\end{subfigure}
\end{figure*}

\begin{figure*}
\centering
\ContinuedFloat 
\begin{subfigure}[b]{\textwidth}
\centering
\includegraphics[width=0.9\textwidth]{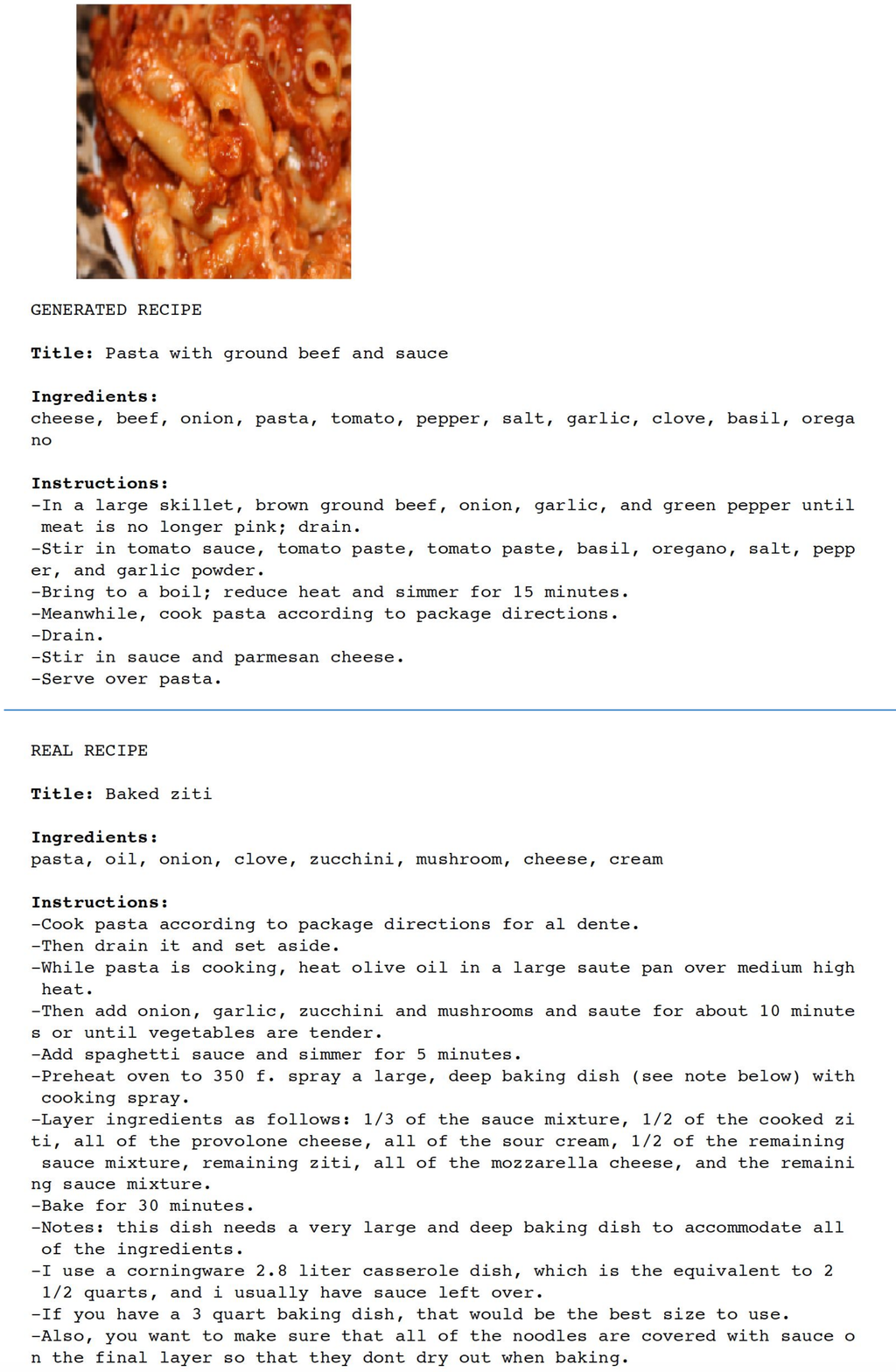}
\caption{\label{fig:recipe_3}}
\end{subfigure}
\end{figure*}

\begin{figure*}
\centering
\ContinuedFloat 
\begin{subfigure}[b]{\textwidth}
\centering
\includegraphics[trim={0 200 0 0},clip,width=\textwidth]{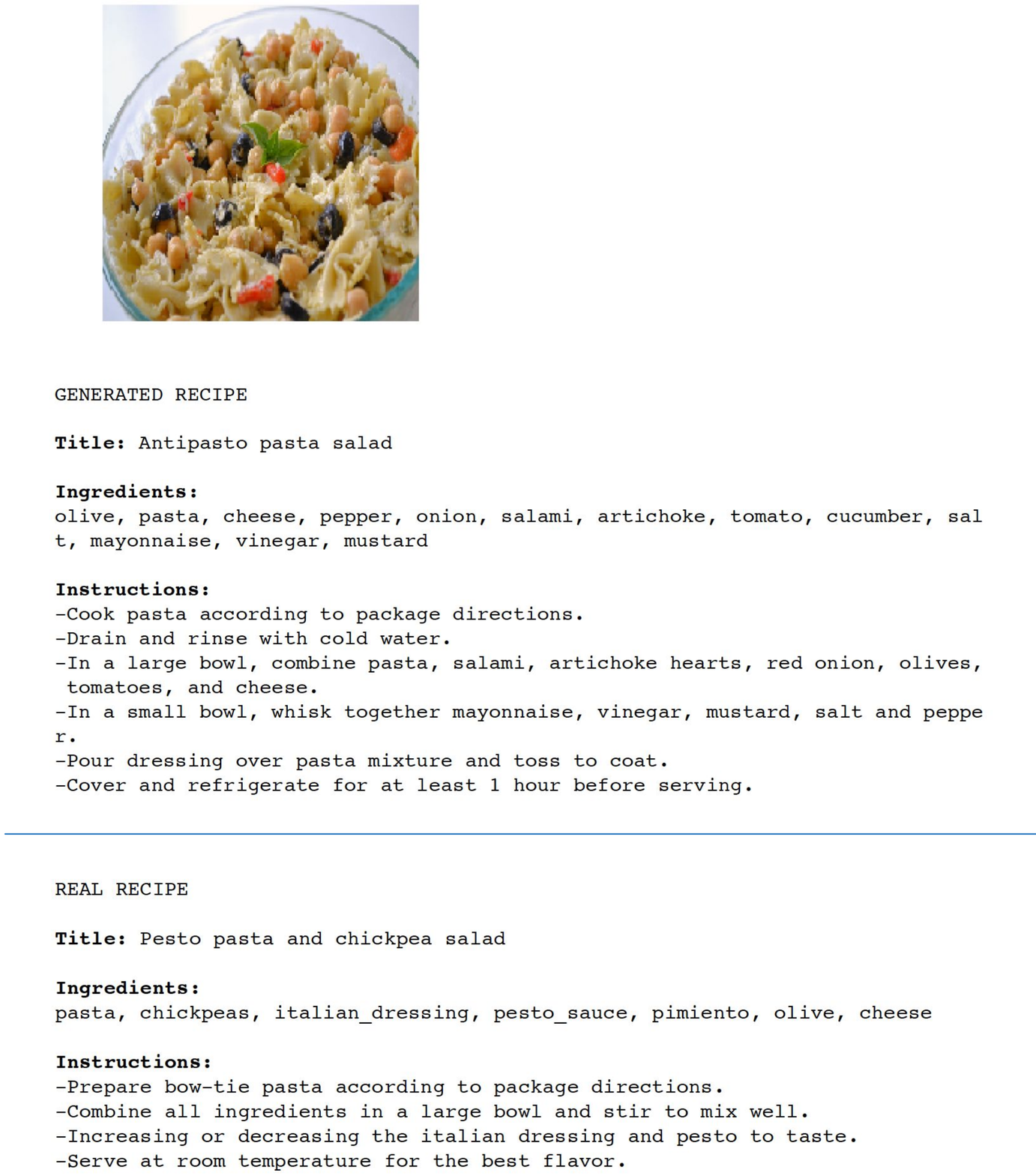}
\caption{\label{fig:recipe_5}}
\end{subfigure}
\end{figure*}

\begin{figure*}
\centering
\ContinuedFloat 
\begin{subfigure}[b]{\textwidth}
\centering
\includegraphics[trim={0 100 0 0},clip,width=\textwidth]{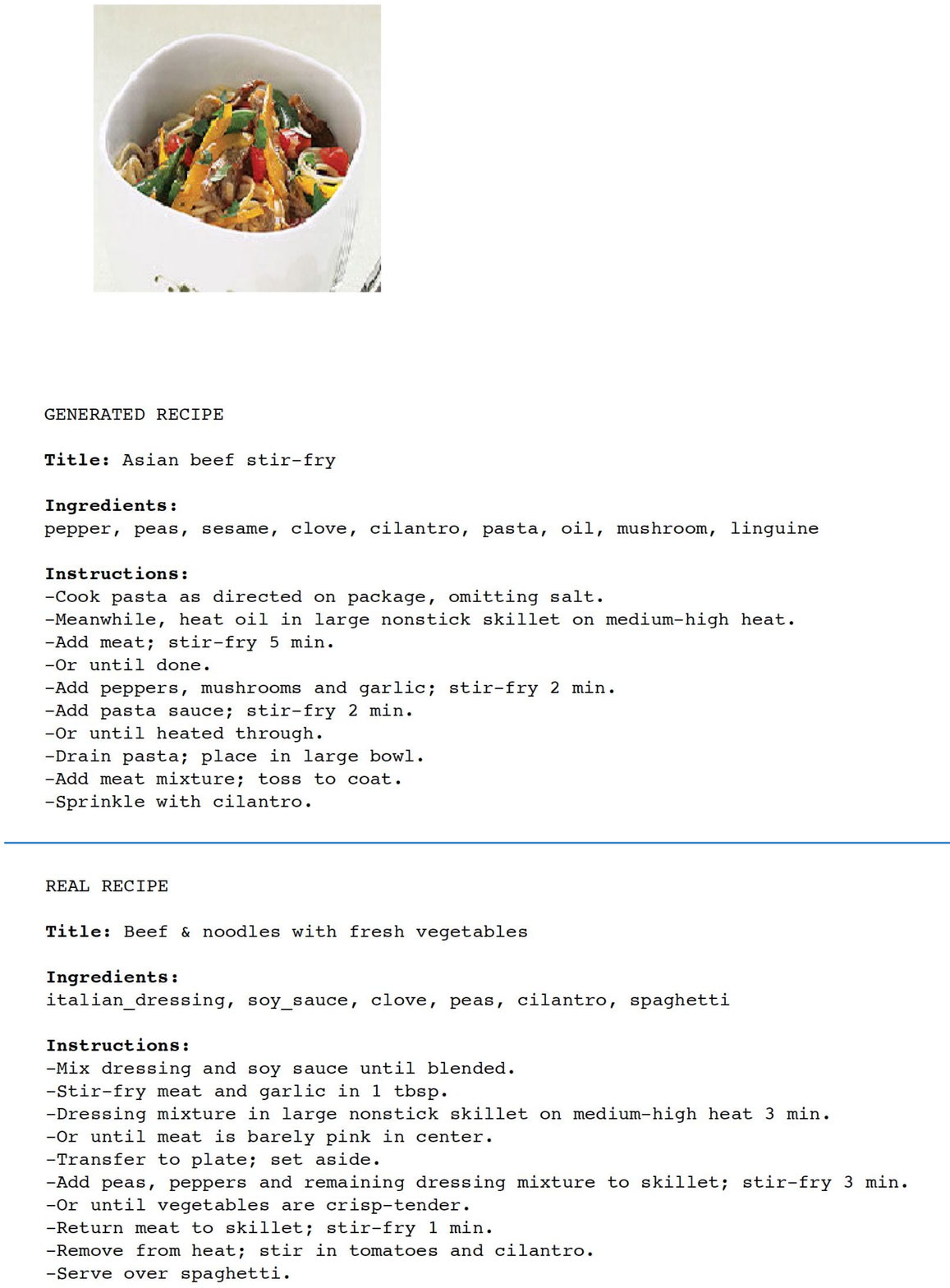}
\caption{\label{fig:recipe_8}}
\end{subfigure}

\end{figure*}

\begin{figure*}
\centering
\ContinuedFloat
\begin{subfigure}[b]{\textwidth}
\centering
\includegraphics[trim={0 200 0 0},clip,width=\textwidth]{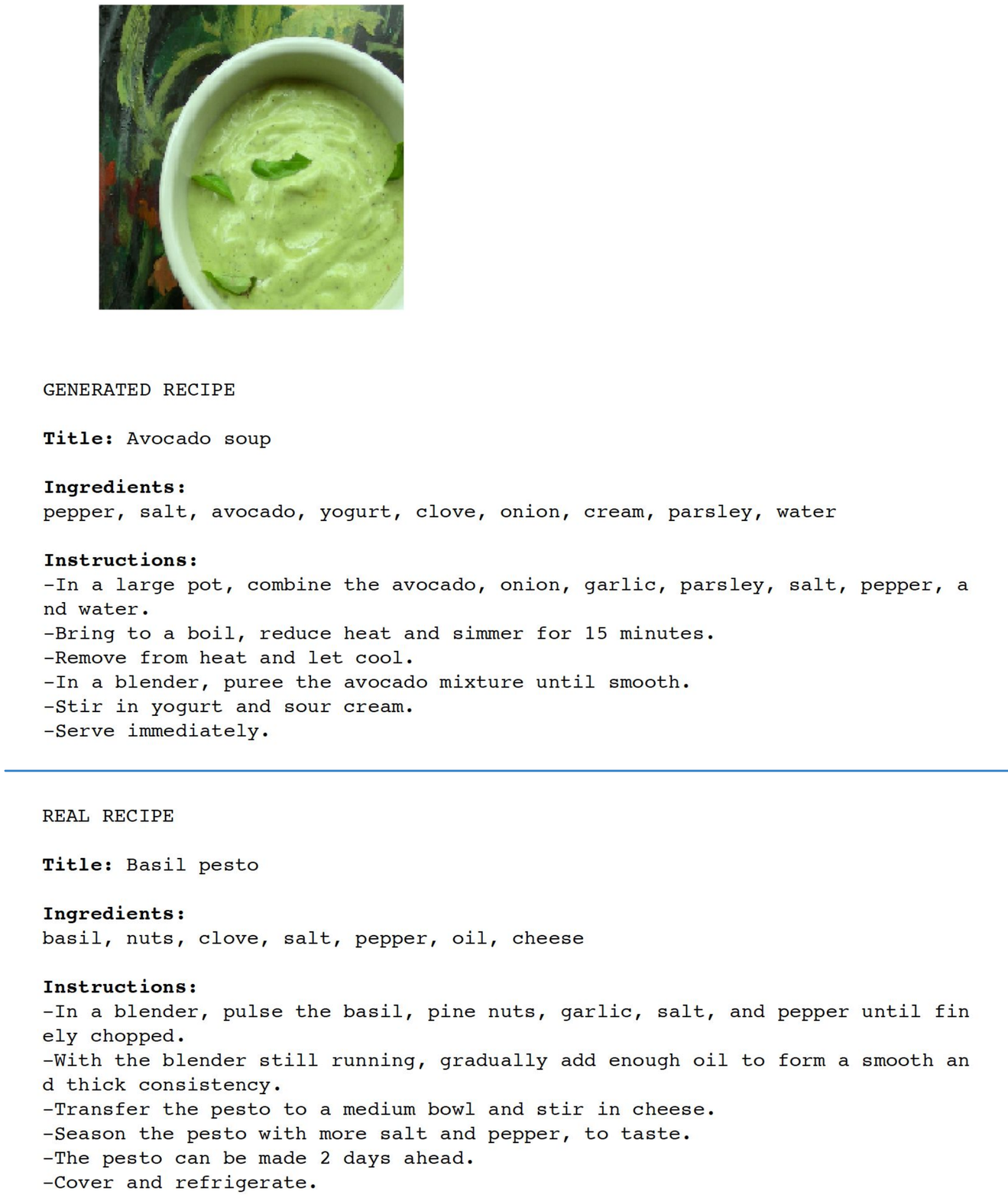}
\caption{\label{fig:recipe_10}}
\end{subfigure}
\end{figure*}

\subsection{Dine Out: A case study}
\label{sec:sm_dineout} 

\begin{figure*}
\centering
\caption{\textbf{Dine Out Study.} Generated recipes for food images taken by authors.}
\label{fig:sm_dineout}
\begin{subfigure}[b]{\textwidth}
\centering
\includegraphics[trim={0 0 0 0},clip,width=\textwidth]{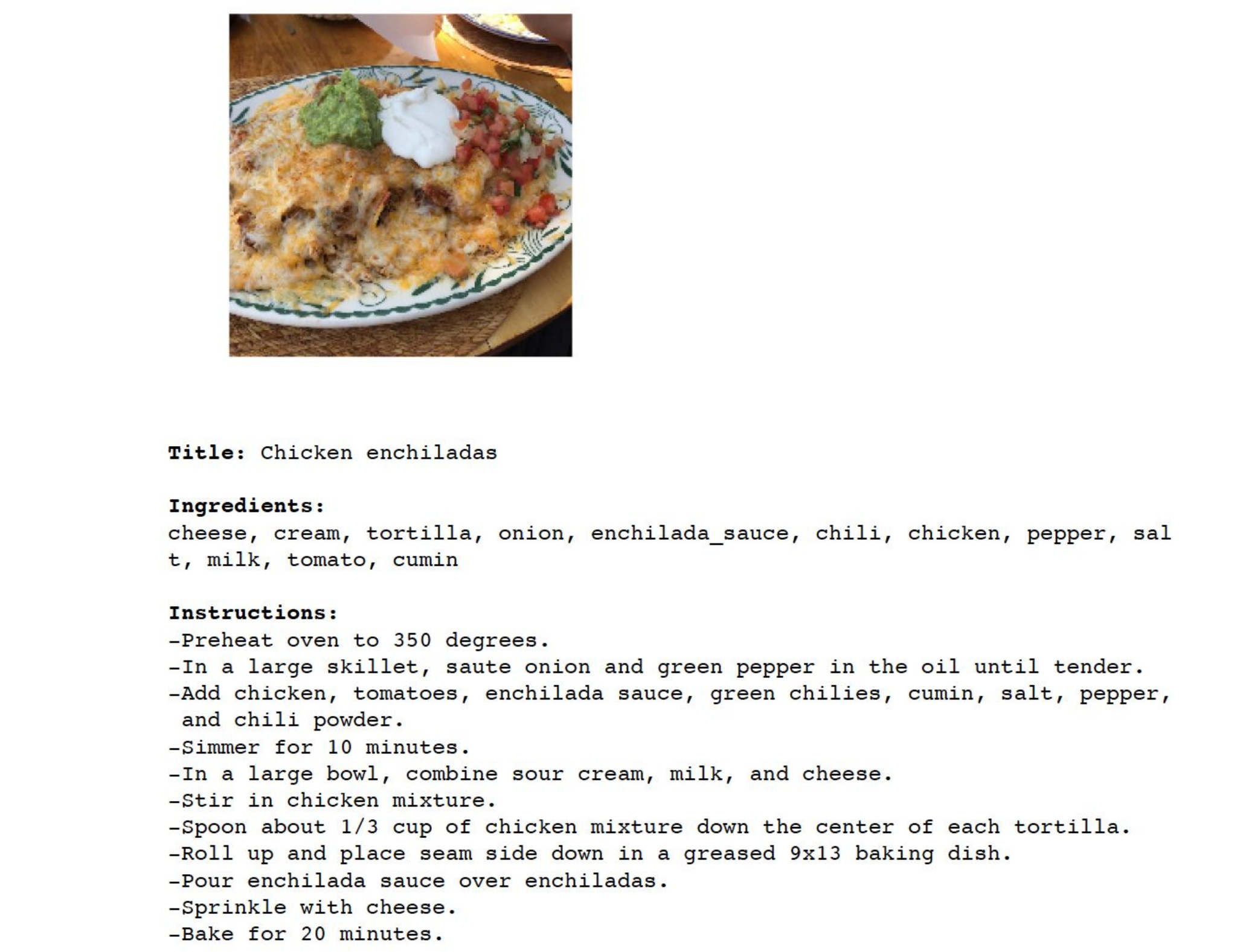}
\caption{\label{fig:do_1}}
\end{subfigure}
\end{figure*}

\begin{figure*}
\centering
\ContinuedFloat 
\begin{subfigure}[b]{\textwidth}
\centering
\includegraphics[trim={0 100 0 0},clip,width=\textwidth]{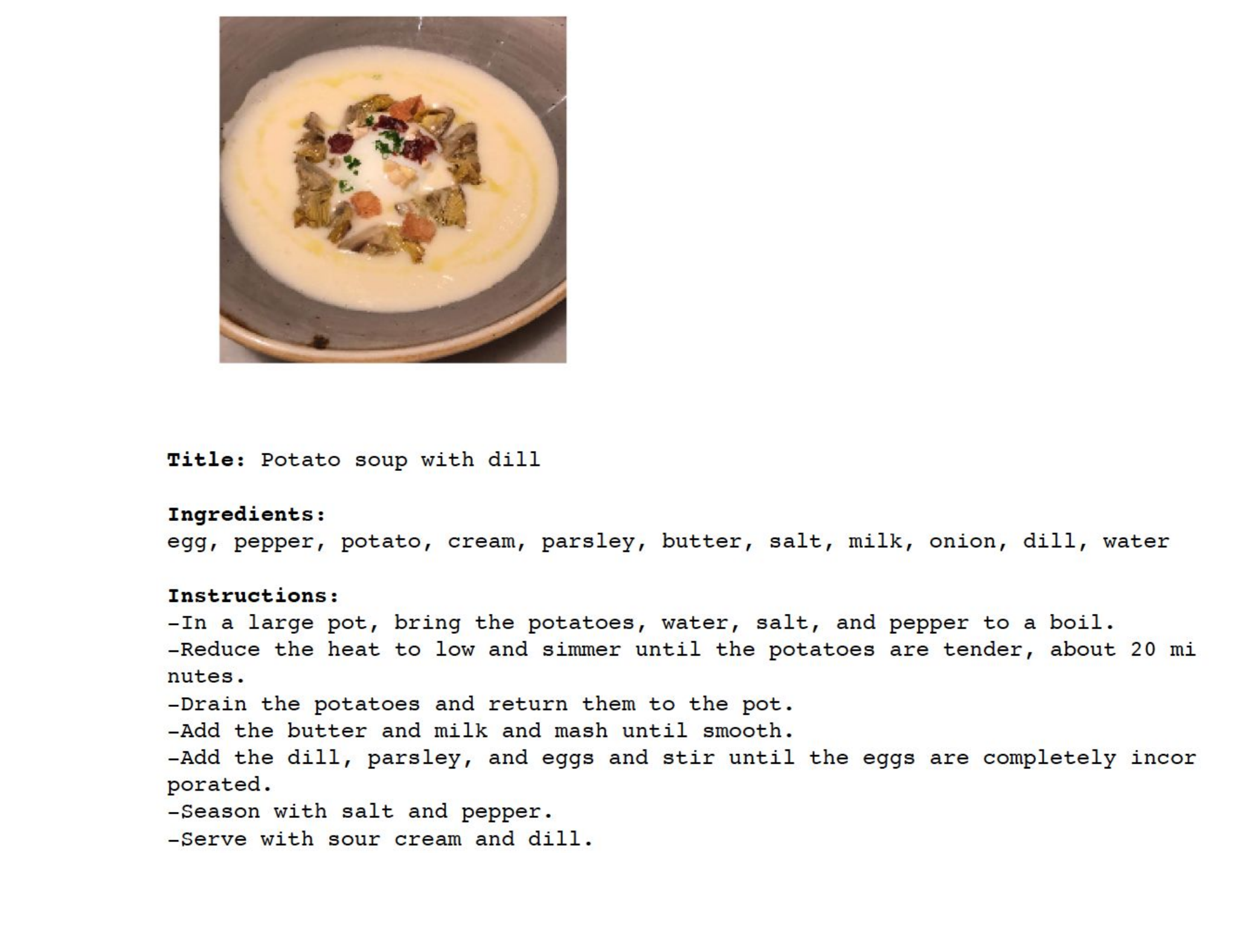}
\caption{\label{fig:do_4}}
\end{subfigure}
\end{figure*}

\begin{figure*}
\centering
\ContinuedFloat 
\begin{subfigure}[b]{\textwidth}
\centering
\includegraphics[trim={0 100 0 0},clip,width=\textwidth]{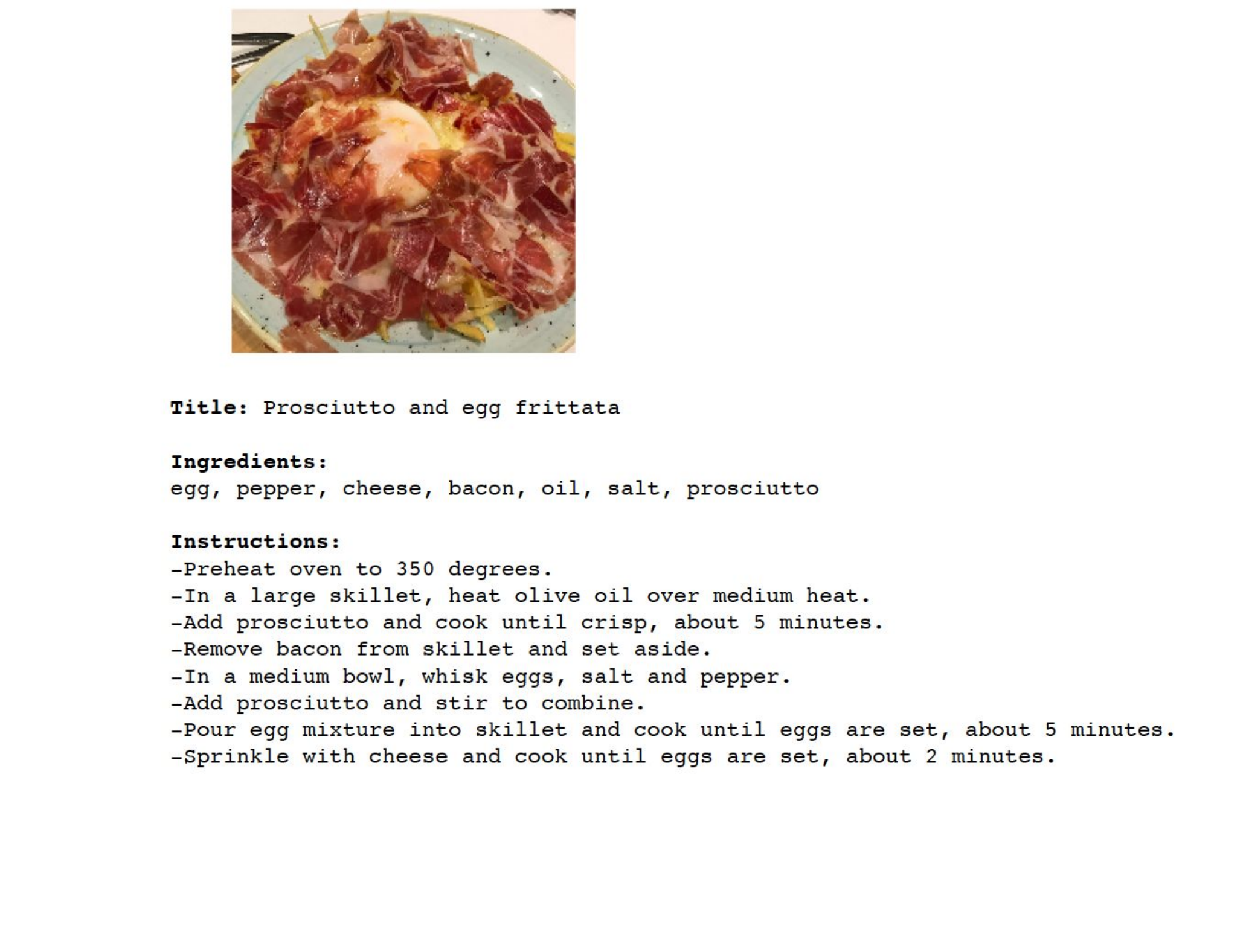}
\caption{\label{fig:do_2}}
\end{subfigure}
\end{figure*}

\begin{figure*}
\centering
\ContinuedFloat 
\begin{subfigure}[b]{\textwidth}
\centering
\includegraphics[trim={0 75 0 20},clip,width=\textwidth]{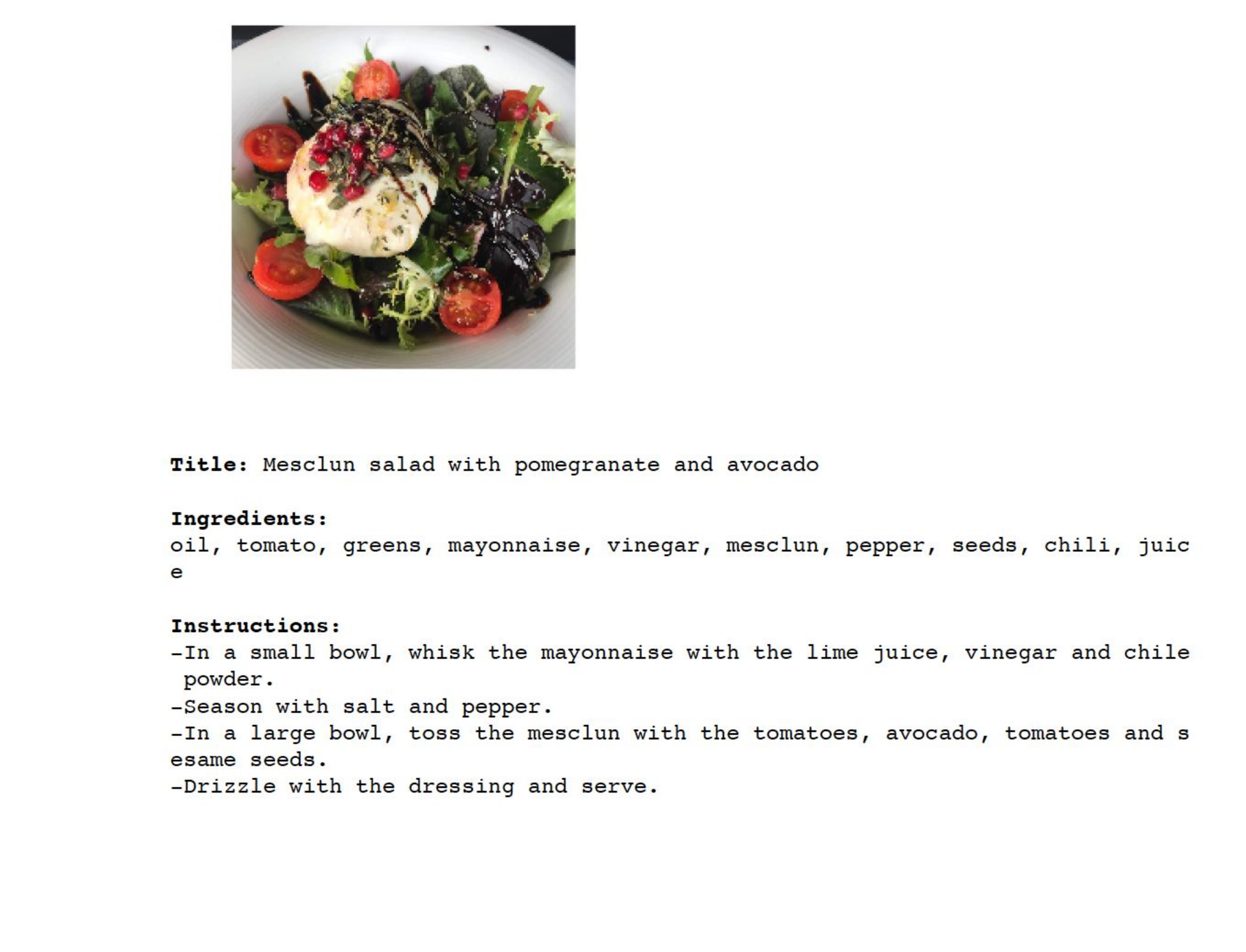}
\caption{\label{fig:do_5}}
\end{subfigure}
\end{figure*}

We test the capabilities of our model to generalize for out-of-dataset food images. Figure \ref{fig:sm_dineout} shows recipes obtained for food images taken by authors at their homes or in restaurants during the weeks prior to the submission.

\end{document}